\documentclass[twocolumn,letterpaper]{IEEEAerospaceCLS}

\usepackage{amsmath,amssymb,amsfonts,mathtools,physics,bm,bbm,dsfont,siunitx}
\AtBeginDocument{\RenewCommandCopy\qty\SI}

\usepackage{booktabs}    
\usepackage{multirow}    
\usepackage{multicol}    
\usepackage{tabularx}    
\usepackage{dcolumn}     

\usepackage{graphicx}
\usepackage{subcaption}
\usepackage{float}
\usepackage{dblfloatfix}

\usepackage{xcolor}
\usepackage{textcomp}

\usepackage{algorithmic}

\usepackage[colorlinks=true, urlcolor=blue, linkcolor=black, citecolor=black]{hyperref}
\usepackage{cleveref}

\usepackage{cite}
\usepackage{url}
\usepackage{IEEEtrantools}

\newcommand{\ignore}[1]{}
\graphicspath{{figures/}}

\makeatletter
\def\itemize{
    \tmpitemindent\itemindent
    \ifnum \@itemdepth > 3 \@toodeep\else
        \advance\@itemdepth\@ne
        \edef\@itemitem{labelitem\romannumeral\the\@itemdepth}
        \list{\csname\@itemitem\endcsname}{
            \itemindent\tmpitemindent
            \itemsep=0pt  
            \parsep=5pt   
            \def\makelabel##1{\hspace\labelsep\hfil{##1}}
        }
    \fi
}
\makeatother

\crefname{figure}{Fig.}{Figs.}
\Crefname{figure}{Figure}{Figures}
\crefname{equation}{Eq.}{Eqs.}
\Crefname{equation}{Equation}{Equations}
\crefname{table}{Table}{Tables}
\Crefname{table}{Table}{Tables}
\crefname{section}{Section}{Sections}
\Crefname{section}{Section}{Sections}

\begin{document}
\bstctlcite{IEEEexample:BSTcontrol}

\title{Semantic Segmentation and Depth Estimation for Real-Time Lunar Surface Mapping Using 3D Gaussian Splatting}

\author{%
    Guillem Casadesus Vila, Adam Dai, and Grace Gao\\
    Department of Aeronautics and Astronautics\\
    Stanford University\\
    Stanford, CA, United States\\
    \{guillemc,adai,gracegao\}@stanford.edu
}

\maketitle

\thispagestyle{plain}
\pagestyle{plain}

\begin{abstract}
	Navigation and mapping on the lunar surface require robust perception under challenging conditions, including poorly textured environments, high-contrast lighting, and limited computational resources. This paper presents a real-time mapping framework that integrates dense perception models with a 3D Gaussian Splatting (3DGS) representation. We first benchmark several models on synthetic datasets generated with the LuPNT simulator, selecting a stereo dense depth estimation model based on Gated Recurrent Units for its balance of speed and accuracy in depth estimation, and a convolutional neural network for its superior performance in detecting semantic segments. Using ground truth poses to decouple the local scene understanding from the global state estimation, our pipeline reconstructs a 120-meter traverse with a geometric height accuracy of approximately 3 cm, outperforming a traditional point cloud baseline without LiDAR. The resulting 3DGS map enables novel view synthesis and serves as a foundation for a full SLAM system, where its capacity for joint map and pose optimization would offer significant advantages. Our results demonstrate that combining semantic segmentation and dense depth estimation with learned map representations is an effective approach for creating detailed, large-scale maps to support future lunar surface missions.
\end{abstract}
\tableofcontents
\section{Introduction}

\subsection{Motivation}
\begin{figure}[b]
	\vspace{-1em}
	\centering
	\begin{subfigure}[b]{0.55\linewidth}
		\includegraphics[width=\linewidth,trim=0 0.5 0.5em 0,clip]{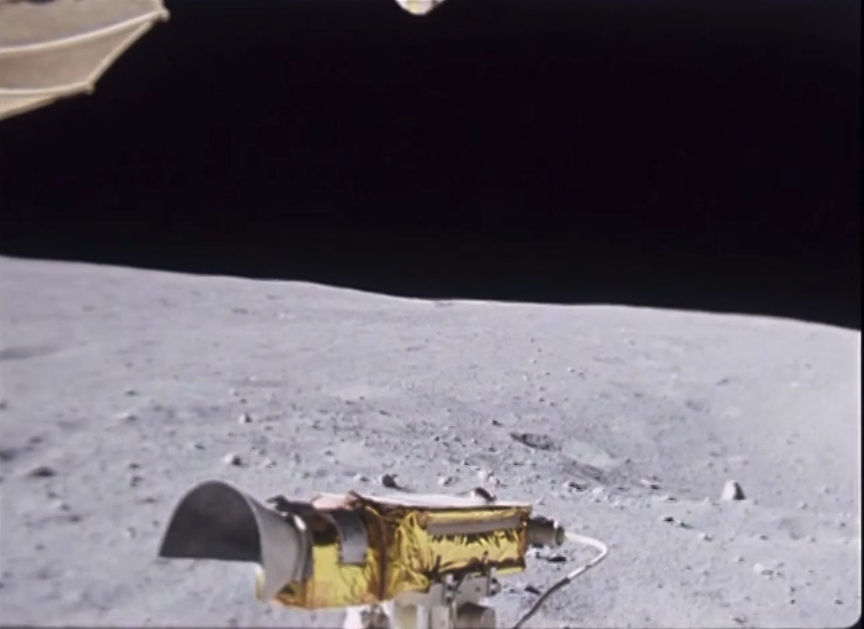}
	\end{subfigure}
	\begin{subfigure}[b]{0.395\linewidth}
		\includegraphics[width=\linewidth,trim=0 0.5 0.5em 0,clip]{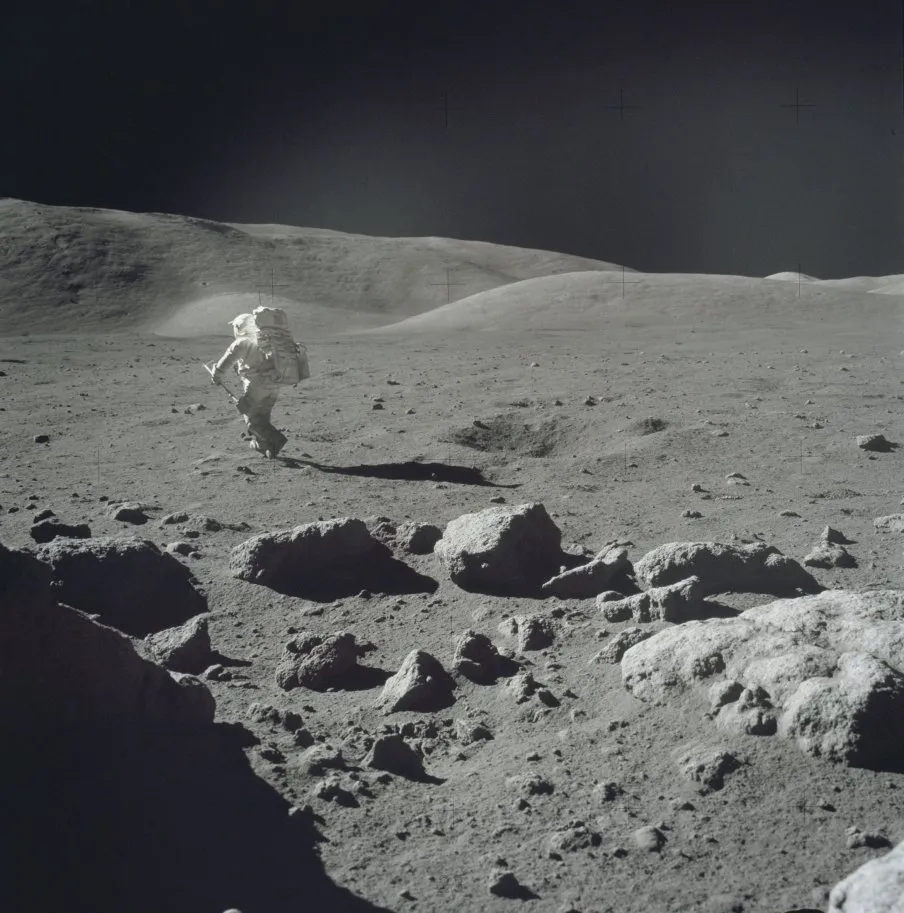}
	\end{subfigure}
	\caption{\bfseries Images from the Apollo missions~\cite{nasa_apollo_2017}.}
	\label{fig:sample_apollo}
\end{figure}

The renewed focus on robotic exploration of the lunar surface brings new challenges for autonomous navigation and mapping~\cite{euroconsult_prospects_2023}. To complete mission objectives, long-range traverses are required in poorly textured environments, and high-contrast and dynamic lighting conditions, as shown in \cref{fig:sample_apollo}. Furthermore, rovers are often limited to radiation-hardened hardware and low-power sensors, prioritizing cameras over 3D active sensor alternatives, like LiDAR. These constraints require the development of computationally efficient, vision-based mapping and localization solutions.

Simultaneous Localization and Mapping (SLAM) is a fundamental technique for an agent to construct a map of an unknown environment while concurrently tracking its position within that map in real-time. In the context of planetary exploration, ``real-time'' implies processing sensor data onboard at a rate sufficient to support continuous rover motion and hazard avoidance. This typically necessitates high-frequency updates for the tracking front-end (e.g., 10 Hz) to ensure stable state estimation, while the mapping back-end can generally operate at a lower rate (e.g., 0.1 to 1 Hz) to incrementally refine a global map.

Classical SLAM methods, such as those based on geometric features~\cite{campos_orb-slam3_2021,newcombe_kinectfusion_2011}, can operate in real time but exhibit reduced performance in poorly textured and high-contrast lighting conditions. While learning-based SLAM methods can improve robustness by learning feature representations from data, their generalization to novel environments like the lunar surface is not guaranteed. A common limitation of both classical and early learning-based methods is their reliance on discrete map representations (e.g., point clouds, voxels), which can be memory-intensive and may not capture fine surface detail.

Recent advances in neural scene representations have produced methods capable of high-fidelity 3D reconstruction from images. For terrestrial applications, recent work has proposed a variety of approaches using different sensor inputs (e.g., monocular, stereo, depth, or IMU) and scene encodings such as Neural Radiance Fields (NeRF) and 3D Gaussian Splatting (3DGS)~\cite{tosi_how_2024}. These systems often rely on perception models for tasks like feature extraction, depth estimation, or semantic understanding. However, since these perception components are typically trained on terrestrial data, it is not clear whether their performance will generalize to the unique visual characteristics of extraterrestrial environments.

\subsection{Related Work}

Semantic characterization of extraterrestrial scenes is an active area of research. A systematic literature review on semantic terrain segmentation for planetary rovers~\cite{kuang_semantic_2022} highlights a gap in existing solutions, noting that no current method simultaneously achieves pixel-level accuracy, real-time inference, and compatibility with onboard hardware. Specific models have shown promise for feature identification, such as detecting impact craters from digital elevation models~\cite{jia_moon_2021}. Despite these advances in perception, the integration of such semantic information into modern neural mapping frameworks for planetary surfaces has not been thoroughly investigated.

Several works have applied neural scene representations to model the lunar surface. Some have focused on surface reconstruction from orbital imagery to generate digital elevation models, often to handle challenging illumination in permanently shadowed regions~\cite{van_kints_neural_2025, adams_summary_2023}. More relevant to rover navigation, a few recent studies have explored using NeRFs with surface-level imagery for localization, mapping, and path planning~\cite{huang_monocular_2025, hansen_analyzing_2024, dai_neural_2023, zhang_neural_2024}.

While these approaches demonstrate the potential of neural representations for capturing lunar terrain, they mostly rely on NeRFs~\cite{mildenhall_nerf_2021}, which have several limitations for real-time mapping. First, NeRFs' rendering process is computationally expensive due to its reliance on volumetric sampling, making it ill-suited for real-time applications on resource-constrained hardware. Second, the underlying multi-layer perceptron (MLP) represents a fixed volume, which is difficult to extend as the rover explores new areas and is not easily deformable to accommodate loop closures. This necessitates stitching multiple models, which can introduce inconsistencies. Finally, the implicit nature of NeRFs makes it less interpretable and difficult to integrate with or query for explicit semantic information. Many of these works require that the camera viewpoints are thoroughly spread out over the scene to constrain the 3D geometry. With typical long rover traverses, this requirement of varied viewpoints would often impede other mission objectives.

\subsection{Contributions}
This work addresses the aforementioned limitations by developing a framework for building semantic, large-scale maps of the lunar surface in real time. Our contributions are:
\begin{itemize}
	\item We propose a real-time mapping framework for lunar surface environments based on 3DGS, which offers fast rendering and a flexible, explicit scene representation suitable for incremental updates.
	\item We integrate and benchmark several perception models, including semantic segmentation and both stereo and monocular depth estimation networks.
	\item We analyze the direct impact of these perception inputs on the geometric and semantic quality of the final 3D reconstruction.
\end{itemize}

\subsection{Paper Organization}
The rest of the paper is organized as follows. \Cref{sec:datasets_and_models} describes the selected dataset and details the perception models used for monocular depth estimation, stereo depth estimation, and semantic segmentation, as well as the fundamentals of 3DGS. \Cref{sec:methodology} presents our proposed real-time mapping framework, outlining the perception front-end, the incremental mapping process, and the optimization back-end, including the loss function. \Cref{sec:results} defines the evaluation metrics and presents a quantitative and qualitative analysis of the dense depth estimation, semantic segmentation, and final surface reconstruction results. Finally, \Cref{sec:conclusion} concludes the paper with a summary of our findings and a discussion of future work.



\section{Datasets and Models}
\label{sec:datasets_and_models}
This section details the datasets and models that are used in our mapping framework. We first introduce the synthetic lunar datasets generated with LuPNT~\cite{vila_lupnt_2025}, an open-source simulator for astrodynamics, communication, and Positioning, Navigation, and Timing (PNT). Then, we provide a comprehensive overview of various methods for two key perception tasks: dense depth estimation—from both monocular and stereo inputs—and semantic segmentation. The primary objective is to benchmark these models to characterize their performance within a lunar context and determine their suitability for the downstream task of real-time 3D reconstruction. Finally, we describe the fundamentals of 3D Gaussian Splatting, the scene representation used in our mapping pipeline.

\subsection{Datasets}
\begin{figure}[b!]
	\centering
	\includegraphics[width=\linewidth, trim=0 1em 0 1em, clip]{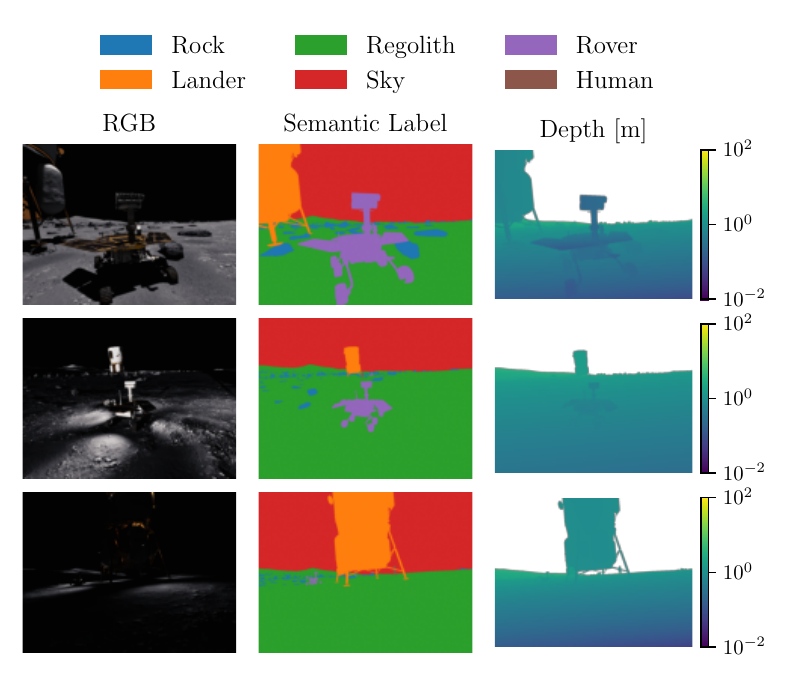}
	\caption{\bfseries Spirals dataset samples.}
	\label{fig:dataset}
\end{figure}
Existing datasets~\cite{liu_lusnarlunar_2024} typically provide static data sequences that do not allow controlling lighting conditions and camera effects. To address this, the datasets used in this work are generated with LuPNT~\cite{vila_lupnt_2025}. The simulator is written in C++ with Python bindings, and uses Unreal Engine to simulate the lunar environment and generate stereo images, semantic labels, dense depth maps, and LiDAR point clouds from camera poses or rover trajectories. The simulated terrain is based on lunar south pole Digital Elevation Models (DEMs) with procedurally generated rocks and craters. The simulator supports multiple agents and allows configuring the Sun position, camera effects, and additional light sources, as well as closed-loop rover control. We use two different datasets:
\begin{itemize}
	\item \textbf{Spirals Dataset:} contains 14,400 stereo image pairs with depth and semantic labels. The images were collected along spiral paths around rovers, landers, and astronauts. This dataset includes variations in camera height, Sun position, and lighting conditions, as shown in \cref{fig:dataset}. It is split into training, evaluation, and test sets.
	\item \textbf{Trajectories Dataset:} contains rover trajectories around a lander, each with 4,300 stereo image pairs captured at 10 Hz. It includes trajectories with different camera effects, low Sun elevation, and rover lights.
\end{itemize}

\subsection{Stereo Depth Estimation}
\begin{figure}[t]
	\centering
	\includegraphics[width=\linewidth]{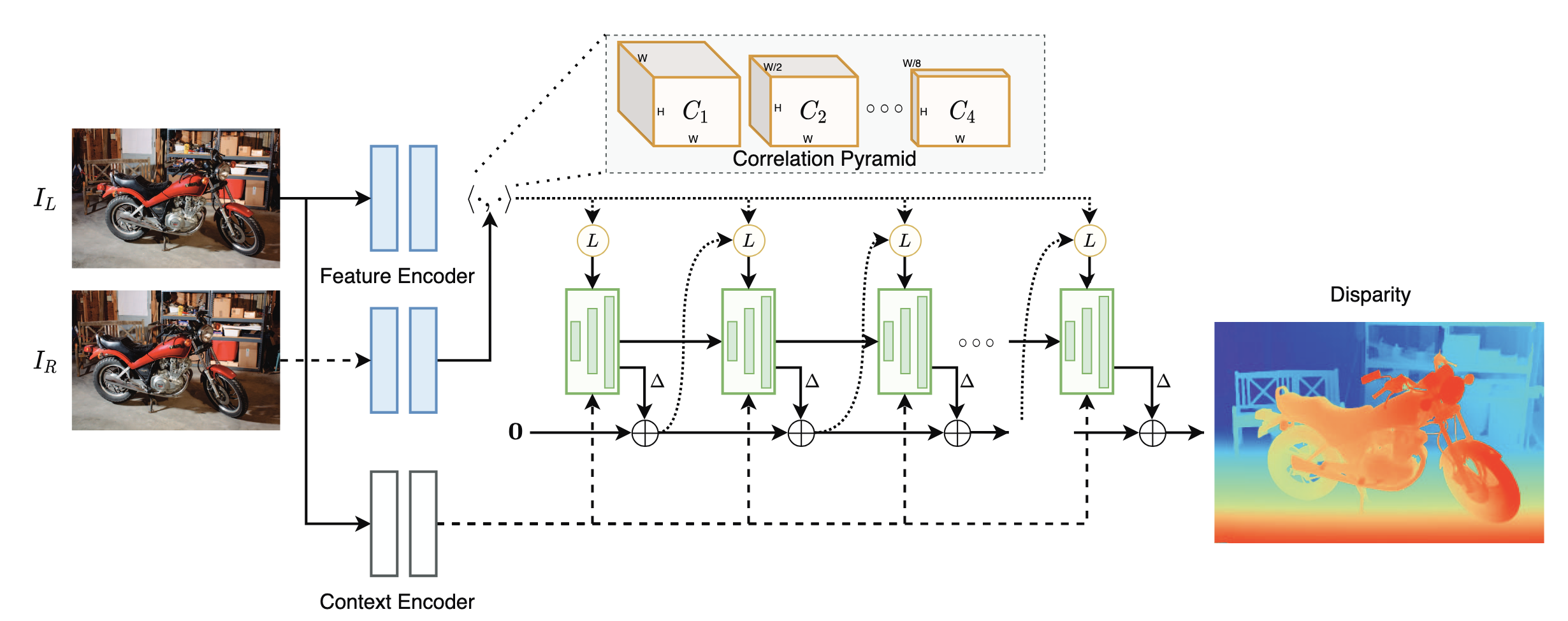}
	\caption{\bfseries RAFT-Stereo architecture (adapted from \cite{lipson_raft-stereo_2021}).}
	\label{fig:raft_stereo}
\end{figure}

We evaluate the following stereo depth estimation models:
\label{sec:stereo_depth_estimation}
\begin{itemize}
	\item \textbf{Block Matching (BM)}: Computes disparity by comparing fixed-size image patches along epipolar lines using a cost function, such as Sum of Absolute Differences (SAD). Its performance is limited in textureless regions and under non-ideal illumination.

	\item \textbf{Semi-Global Matching (SGM)}~\cite{hirschmuller_stereo_2008}: Aggregates matching costs along multiple 1D paths across the image to approximate a 2D global smoothness constraint. This approach combines the efficiency of local methods with improved performance in low-texture areas.

	\item \textbf{RAFT-Stereo}~\cite{lipson_raft-stereo_2021}: Shown in \cref{fig:raft_stereo}, it adapts the RAFT architecture from optical flow by constructing a 3D correlation volume of all disparities at all pixels. A recurrent, GRU-based unit then iteratively updates a high-resolution disparity field from this volume.

	\item \textbf{CREStereo}~\cite{li_practical_2022}: A cascaded recurrent network that operates in a coarse-to-fine manner. It uses recurrent refinement units at each stage and an adaptive group correlation layer to handle large displacements between stereo images.
\end{itemize}

\subsection{Monocular Depth Estimation}
We evaluate the following monocular dense depth estimation models:
\label{sec:monocular_depth_estimation}
\begin{itemize}
	\item \textbf{Depth Anything V2}~\cite{yang_depth_2024}: A transformer-based model trained on a dataset of over 62 million synthetic and pseudo-labeled real images. It is released in variants from 25M to 1.3B parameters for zero-shot relative depth estimation.

	\item \textbf{GLPN}~\cite{kim_global-local_2022}: An architecture that uses a transformer-based encoder to capture global context and a lightweight decoder. A selective feature fusion module combines features from different stages of the encoder.

	\item \textbf{DPT}~\cite{ranftl_vision_2021}: Uses a Vision Transformer (ViT) as its backbone for dense prediction tasks. It processes feature maps at a constant resolution, providing global receptive fields at every stage of the network.

	\item \textbf{Depth Pro}~\cite{bochkovskii_depth_2025}: A model for zero-shot metric depth estimation that uses a multi-scale vision transformer and is designed to operate on high-resolution inputs.
\end{itemize}

\subsection{Semantic Segmentation}
We evaluate the following semantic segmentation models:
\label{sec:semantic_segmentation}
\begin{itemize}
	\item \textbf{U-Net}~\cite{ronneberger_u-net_2015}: An encoder-decoder architecture with skip connections that concatenate features from the downsampling path to the upsampling path to retain spatial information.

	\item \textbf{U-Net++}~\cite{zhou_unet_2018}: Modifies the U-Net skip pathways with nested and dense convolutional blocks to reduce the semantic gap between encoder and decoder features.

	\item \textbf{MANet}~\cite{fan_ma-net_2020}: Augments an encoder-decoder network with a module that combines position-wise and channel-wise attention to capture contextual dependencies.

	\item \textbf{LinkNet}~\cite{chaurasia_linknet_2017}: A lightweight encoder-decoder network for real-time applications that passes encoder features at each level directly to the corresponding decoder level.

	\item \textbf{FPN}~\cite{lin_feature_2017}: Constructs a multi-scale feature pyramid using a top-down pathway and lateral connections to merge semantic information from deep layers with spatial information from shallow layers.

	\item \textbf{PSPNet}~\cite{zhao_pyramid_2017}: Introduces a pyramid pooling module that applies pooling operations at multiple scales to aggregate global context information.

	\item \textbf{PAN (Path Aggregation Network)}~\cite{li_pyramid_2018}: Augments the top-down feature pyramid with an additional bottom-up pathway to shorten the information path for low-level features.

	\item \textbf{DeepLabV3}~\cite{chen_rethinking_2017}: Uses atrous (dilated) convolutions to control the spatial resolution of feature maps and an Atrous Spatial Pyramid Pooling (ASPP) module to probe features at multiple scales.

	\item \textbf{DeepLabV3+}~\cite{chen_encoder-decoder_2018}: Extends DeepLabV3 by adding a decoder module to refine object boundaries and uses depthwise separable convolutions for computational efficiency.

	\item \textbf{UPerNet}~\cite{xiao_unified_2018}: A unified framework that combines a Feature Pyramid Network (FPN) backbone with a pyramid pooling module to parse features at various scales simultaneously.

	\item \textbf{Segformer}~\cite{xie_segformer_2021}: A transformer-based model using a hierarchical transformer encoder to produce multi-scale features without positional encodings, combined with a lightweight multilayer perceptron (MLP) decoder.

	\item \textbf{DPT}~\cite{ranftl_vision_2021}: Uses a Vision Transformer (ViT) as its backbone, providing global receptive fields at every stage of the feature extraction process for dense prediction tasks.
\end{itemize}

\subsection{3D Gaussian Splatting}
\begin{figure}[t]
	\centering
	\includegraphics[width=\linewidth]{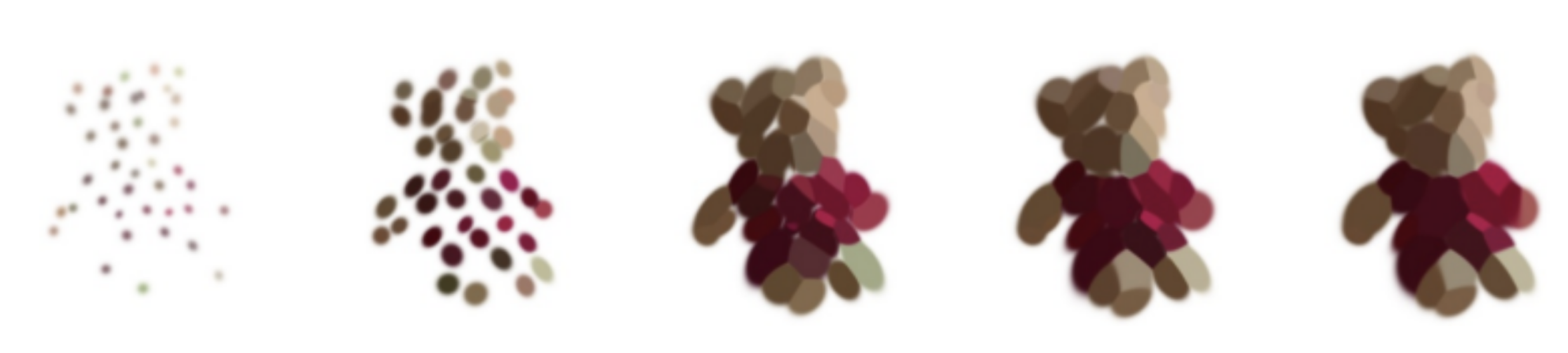}
	\caption{\bfseries 3D Gaussian Splatting (adapted from \cite{dalal_gaussian_2024}).}
	\label{fig:gaussian_splatting}
\end{figure}

3D Gaussian Splatting (3DGS)~\cite{kerbl_3d_2023} is a rasterization-based method for novel view synthesis that represents a 3D scene with a collection of explicit, optimizable primitives. Unlike implicit representations like NeRF~\cite{mildenhall_nerf_2021}, 3DGS uses thousands to millions of 3D Gaussians to explicitly model the scene's geometry and appearance, as illustrated in \cref{fig:gaussian_splatting}.
Each Gaussian is defined by a set of learnable parameters:
\begin{itemize}
	\item \textbf{Mean:} $\boldsymbol{\mu} \in \mathbb{R}^3$ determines its location in 3D space.
	\item \textbf{Covariance:} $\boldsymbol{\Sigma} \in \mathbb{R}^{3 \times 3}$ defines its shape and orientation. To ensure $\boldsymbol{\Sigma}$ is always a valid positive semi-definite matrix and to allow for intuitive optimization, it is parameterized by a scaling vector $\mathbf{s} \in \mathbb{R}^3$ and a rotation quaternion $\mathbf{q} \in \mathbb{R}^4$.
	      \begin{equation}
		      \boldsymbol{\Sigma} = \mathbf{R} \mathbf{S} \mathbf{S}^\top \mathbf{R}^\top
	      \end{equation}
	      where $\mathbf{R}$ is the rotation matrix derived from $\mathbf{q}$ and $\mathbf{S}$ is a diagonal scaling matrix derived from $\mathbf{s}$.
	\item \textbf{Opacity:} $\alpha \in [0, 1]$ controls the transparency of the Gaussian.
	\item \textbf{Color:} View-dependent color is modeled using Spherical Harmonics (SH) coefficients.
\end{itemize}

\subsubsection{Differentiable Rendering}
To synthesize outputs from a novel viewpoint, the 3D Gaussians are projected onto the 2D image plane, forming elliptical splats. These splats are sorted by depth and composited in a front-to-back order using alpha blending. This differentiable rendering process produces not only the final RGB image ($\hat{I}$), but also a depth map and an accumulation (opacity) map. The rendered image can be directly compared to a ground truth image for gradient-based optimization, while the depth and accumulation maps provide additional supervision or regularization signals as needed.

\subsubsection{Adaptive Densification Strategy}
A key component of the 3DGS training process is the strategy for adaptive densification, which dynamically adjusts the set of Gaussians to efficiently represent the scene~\cite{kerbl_3d_2023}. This process is governed by periodic checks that add or remove primitives based on three main heuristics: the magnitude of the positional image-plane gradient, the 3D scale, and the opacity of each Gaussian. The strategy consists of three primary operations: growing, pruning, and opacity reset.

\begin{itemize}
	\item \textbf{Growing (Densification)}:
	      To represent complex regions that are not yet well-reconstructed, new Gaussians are introduced where the image-plane gradients exceed a threshold. This indicates that the optimizer is struggling to place the existing Gaussians correctly. Two methods are used:
	      \begin{itemize}
		      \item Duplication: If a Gaussian has a high gradient and small 3D scale, it is duplicated to add detail in under-reconstructed regions.
		      \item Splitting: If a Gaussian has a high gradient and large 3D scale, it is split into two smaller Gaussians to better capture complex geometry.
	      \end{itemize}

	\item \textbf{Pruning}:
	      To maintain a compact representation and remove artifacts, unnecessary Gaussians are pruned. A Gaussian is removed if it meets certain criteria, such as:
	      \begin{itemize}
		      \item Its opacity $\alpha$ falls below a minimum threshold, rendering it effectively invisible.
		      \item Its 3D scale grows excessively large, which can cause blurry or hazy artifacts.
	      \end{itemize}

	\item \textbf{Opacity Reset}:
	      Periodically during training, the opacities of all Gaussians are reset to a low value. This acts as a regularizer, forcing the model to re-evaluate the importance of each Gaussian. Primitives that are essential to the reconstruction will quickly regain high opacity, while transient or unnecessary ones will fail to do so and be removed in a subsequent pruning phase.
\end{itemize}

\subsubsection{Loss Function}
The total loss $\mathcal{L}_{\text{total}}$ used to train the standard 3DGS model is a weighted sum of a reconstruction loss and a scale regularization term.

The reconstruction loss $\mathcal{L}_{\text{recon}}$ combines the L1 and D-SSIM losses, balanced by a hyperparameter $\lambda_{\text{SSIM}}$:
\begin{equation}
	\mathcal{L}_{\text{recon}} = (1 - \lambda_{\text{SSIM}}) \cdot \|I - \hat{I}\|_1 + \lambda_{\text{SSIM}} \cdot \left(1 - \text{SSIM}(I, \hat{I})\right)
\end{equation}
where $I$ is the ground truth image and $\hat{I}$ is the rendered image.

To prevent the Gaussians from becoming overly stretched, a scale regularization loss $\mathcal{L}_{\text{scale}}$ is applied. For each Gaussian $g$ with a scale vector $\mathbf{s}_g$ out of $N_g$ total Gaussians, this loss penalizes the ratio of its largest to smallest scale component if it exceeds a threshold $\tau_{\text{ratio}}$:
\begin{equation}
	\mathcal{L}_{\text{scale}} = \frac{1}{N_g} \sum_{g=1}^{N_g} \max \left( 0, \frac{\max(\mathbf{s}_g)}{\min(\mathbf{s}_g)} - \tau_{\text{ratio}} \right)
\end{equation}

The final loss is the sum of these two components, with a weighting factor $\lambda_{\text{scale}}$ for the regularization term:
\begin{equation}
	\mathcal{L}_{\text{total}} = \mathcal{L}_{\text{recon}} + \lambda_{\text{scale}}\mathcal{L}_{\text{scale}}.
\end{equation}

\begin{figure*}[t]
	\centering
	\includegraphics[width=\linewidth]{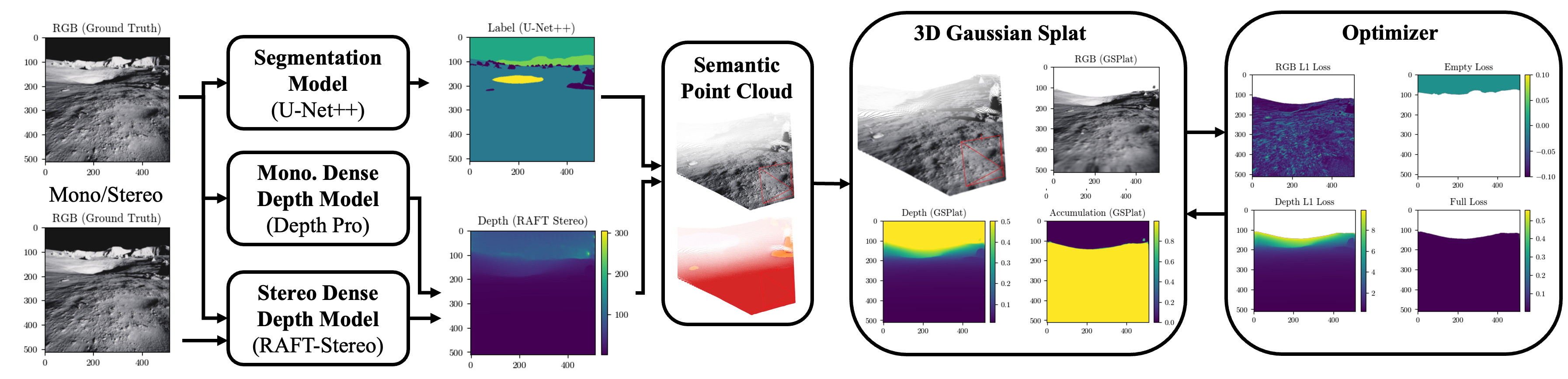}
	\caption{\bfseries Diagram of the proposed real-time 3DGS mapping pipeline.}
	\label{fig:3dgs_diagram}
	\vspace{-0.5em}
\end{figure*}

\vspace{-1.0em}
\section{Methodology}
\label{sec:methodology}
Our primary contribution is a real-time pipeline for the semantic 3D reconstruction of the lunar surface using a 3D Gaussian Splatting (3DGS) representation. The architecture is composed of three principal stages: a perception frontend, an incremental mapping backend, and a keyframe-based optimization engine.

\subsection{Perception and Tracking Frontend}
The frontend processes raw sensor data into a structured format. We assume that rover pose estimates are continuously available from an external tracking system, such as visual-inertial odometry; the development of this tracking component is outside the scope of this work~\cite{dai2025full}. For each incoming camera frame, the frontend computes a dense depth map using either traditional stereo algorithms or monocular depth estimation networks, demonstrating modular support for various sensor configurations. Concurrently, a semantic segmentation network classifies each pixel into relevant lunar categories (e.g., regolith, rocks), providing critical context. Pixels classified as sky are explicitly masked to prevent the erroneous creation of 3D points.

\begin{figure}[t]
	\centering
	\includegraphics[width=\linewidth]{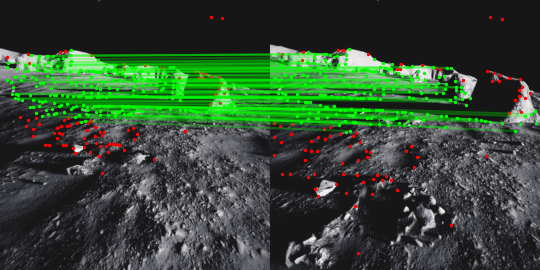}
	\caption{\bfseries Matches between consecutive frames used for monocular depth estimation scaling.~\cite{liu_lusnarlunar_2024}}
	\label{fig:matches}
	\vspace{-1em}
\end{figure}
For monocular depth estimation models, we use a scaling strategy using triangulated points from matched features between consecutive frames. We detect and match keypoints using SuperPoint~\cite{detone_superpoint_2018} and SuperGlue~\cite{sarlin_superglue_2020}, as shown in \cref{fig:matches}, then triangulate these points to obtain metric depth estimates. We use these triangulated depths to scale the output of the monocular models. To ensure robust scaling, we apply depth masking to filter out unreliable matches and use random sample consensus (RANSAC) to remove outliers from the triangulated points. Specifically, we solve for the scale parameter $\theta$ and offset $\gamma$ by minimizing:
\begin{equation}
	\min_{\theta,\gamma} \sum_{p \in M_{\text{sparse}}} \|\theta \hat{D}(p) + \gamma - \hat{D}_s(p)\|_2^2,
\end{equation}
where $\hat{D}$ represents the estimated dense depth from the monocular model, $\hat{D}_s$ is the sparse depth from triangulation, and $M_{\text{sparse}}$ is the mask of sparse depth estimates.

\subsection{Incremental Mapping}
The backend receives the processed frames from the frontend and is responsible for incrementally building the global 3D map. For each new frame, the estimated pose, dense depth map, and semantic labels are fused to generate a registered, per-frame 3D semantic point cloud. This step transforms the 2D perception outputs into a 3D point cloud where every point is associated with a color and a semantic class. This point cloud is then used to add new Gaussians to the global 3DGS model.
Each new Gaussian's size is set proportional to the average distance to its nearest neighbors, ensuring adaptive coverage based on point density.
To maintain real-time performance and prevent unbounded map growth over long trajectories, we use a voxel-based filtering strategy. Only points that fall into previously unobserved regions of space are used to initialize new Gaussians. A key aspect of our approach is that each new Gaussian is initialized with the semantic label from its corresponding point.

\subsection{Optimization Backend}
The optimization engine runs as a background process, continuously refining all Gaussian parameters and camera poses to improve global consistency.

\subsubsection{Keyframe-Based Asynchronous Optimization}
A keyframe buffer stores a history of processed frames, serving as a long-term memory to prevent catastrophic forgetting. The optimization engine runs asynchronously, periodically sampling batches of past views from this buffer to enforce global consistency. This process is decoupled from the frame rate and triggered by time intervals, allowing for the efficient use of idle compute time during rover operations. While this work focuses on optimizing the Gaussian parameters, the keyframe-based approach provides a foundation for future extensions such as camera pose refinement and loop closure.

\subsection{Semantic Supervision}
To incorporate semantic information into the mapping process, we leverage the output of the semantic segmentation network. Each 3D Gaussian is assigned a discrete semantic label $l \in L$, where $L$ is the set of all possible classes. During the rasterization process, in addition to rendering RGB and depth, we also rasterize the semantic probabilities. Specifically, for each pixel, the semantic probability $P(l)$ for a class $l$ is computed by accumulating the opacities of Gaussians belonging to that class, similar to how color is rendered. This allows the 3DGS model to represent a continuous semantic field, which is supervised by the predicted segmentation masks from the frontend.

\subsubsection{Densification Strategy}
To manage the set of Gaussians, we modify the standard 3DGS densification strategy. The periodic, global reset of Gaussian opacities is removed; this is critical for long-term mapping as our optimization prioritizes recent keyframes, and a global reset could cause stable, older portions of the map to be incorrectly pruned, leading to catastrophic forgetting. Furthermore, our implementation omits the standard splitting and cloning operations. Since the backend incrementally adds geometry from dense depth maps with each new keyframe, the large, under-reconstructed regions that typically necessitate splitting are less prevalent. Instead, we manage map size with the voxel-based filtering strategy and by pruning Gaussians that are inconsistent with a known prior Digital Elevation Model (DEM).

\subsubsection{Loss Function}
The optimization process minimizes a total loss function, $\mathcal{L}_{\text{total}}$, which extends the standard 3DGS objective ($\mathcal{L}_{\text{recon}} + \mathcal{L}_{\text{scale}}$). The unique characteristics of lunar environments—namely high-contrast lighting and deep, hard-edged shadows with no atmospheric scattering—make relying on a simple background color insufficient. A deep shadow on the regolith can be photometrically indistinguishable from the black sky, providing ambiguous signals to the optimizer. To address this, we introduce explicit geometric and semantic supervision through additional loss components that use masks derived from dense depth and semantic labels.

These losses are applied using two masks. The dense mask, $M_{\text{dense}}$, identifies pixels corresponding to valid geometry. A pixel is included in this mask only if it is not classified as sky and it is within a pre-specified depth range for optimization.
In this sense, a pixel that corresponds to a rock that is far from the current camera pose is not included in the dense mask.
The empty space mask, $M_{\text{empty}}$, identifies pixels known to contain no geometry and is composed of pixels identified as sky or beyond the depth range for optimization.
Our additional loss components are:
\begin{itemize}
	\item \textbf{Supervised Depth Loss}: To leverage the dense depth maps from our perception frontend, we add a supervised depth loss, $\mathcal{L}_{\text{depth}}$. This term enforces geometric accuracy by penalizing the L1 difference between the rendered depth, $\hat{D}$, and the input depth map, $D$, only on the $M_{\text{dense}}$ mask:

	      \begin{equation}
		      \mathcal{L}_{\text{depth}} = \frac{1}{|M_{\text{dense}}|} \sum_{p \in M_{\text{dense}}} \frac{|D(p) - \hat{D}(p)|}{\max_{q \in M_{\text{dense}}} D(q)}
	      \end{equation}

	\item \textbf{Volumetric Regularization Losses}: To regularize the density distribution and remove artifacts, we use two losses based on the rendered alpha accumulation, $\hat{\alpha}$. An \textit{empty loss}, $\mathcal{L}_{\text{empty}}$, penalizes density in regions defined by $M_{\text{empty}}$. A complementary \textit{dense loss}, $\mathcal{L}_{\text{dense}}$, encourages surfaces within $M_{\text{dense}}$ to be opaque:
	      \begin{align}
		      \mathcal{L}_{\text{empty}} & = \frac{1}{|M_{\text{empty}}|} \sum_{p \in M_{\text{empty}}} \hat{\alpha}(p)^{\alpha_{\text{accum}}}       \\
		      \mathcal{L}_{\text{dense}} & = \frac{1}{|M_{\text{dense}}|} \sum_{p \in M_{\text{dense}}} (1 - \hat{\alpha}(p))^{\alpha_{\text{accum}}}
	      \end{align}
	      where $\alpha_{\text{accum}}$ is a hyperparameter that controls the strength of the regularization.

	\item \textbf{Semantic Loss}: We use a Negative Log Likelihood (NLL) loss to supervise the semantic field. The predicted semantic probability for each pixel $\hat{P}(c)$ is compared against the ground truth label $c$ (from the segmentation network). To ensure numerical stability, the probabilities are clamped before computing the loss:
	      \begin{equation}
		      \mathcal{L}_{\text{semantic}} = \frac{1}{|M_{\text{dense}}|} \sum_{p \in M_{\text{dense}}} - \log(\hat{P}_{c(p)}(p))
	      \end{equation}
	      where $c(p)$ is the class label at pixel $p$ and $M_{\text{dense}}$ is the mask of pixels with valid semantic labels.
\end{itemize}

\begin{figure}[h]
	\centering
	\includegraphics[width=\linewidth]{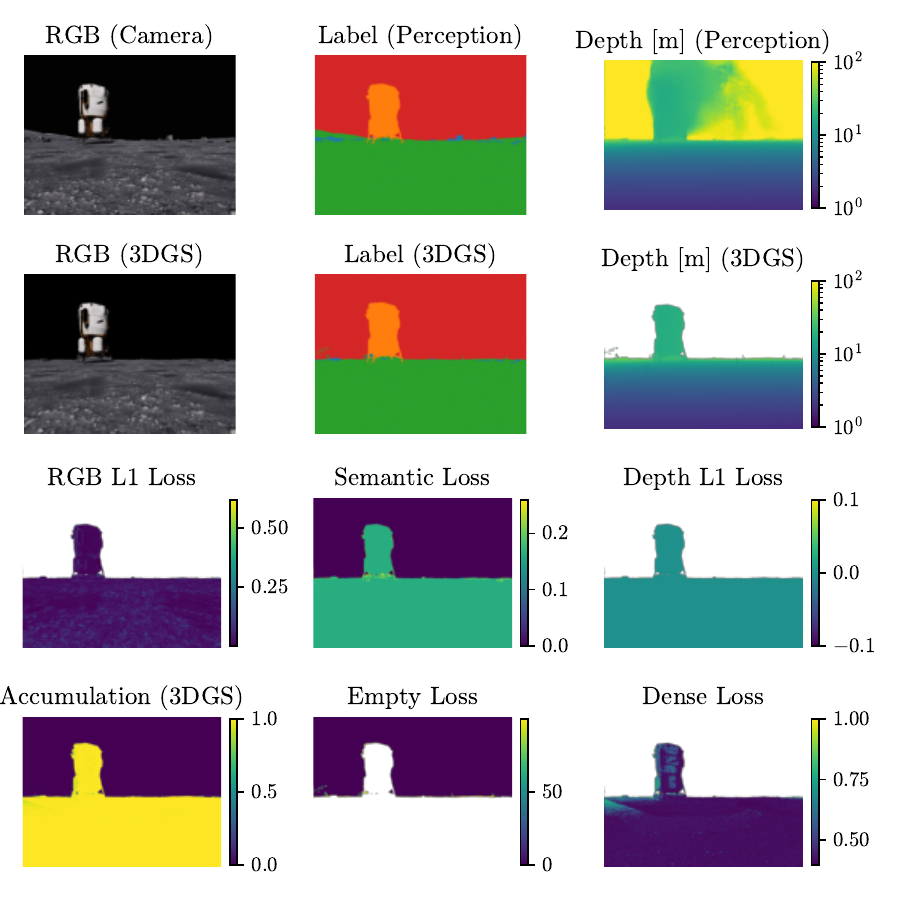}
	\caption{Loss function components.}
	\label{fig:loss}
\end{figure}

The final objective function is a weighted summation of the baseline and our new components:
\begin{align}
	\mathcal{L}_{\text{total}} =\  & \mathcal{L}_{\text{recon}} + \lambda_{\text{scale}}\mathcal{L}_{\text{scale}} + \lambda_{\text{depth}}\mathcal{L}_{\text{depth}} \notag                        \\
	                               & + \lambda_{\text{empty}}\mathcal{L}_{\text{empty}} + \lambda_{\text{dense}}\mathcal{L}_{\text{dense}} + \lambda_{\text{semantic}}\mathcal{L}_{\text{semantic}}
\end{align}
where the $\lambda$ terms are the corresponding weighting coefficients. \Cref{fig:loss} shows an example of the loss function components. The first row includes a camera image from a stereo pair with the predicted dense depth map and semantic segmentation; the second row shows the rendered RGB, depth, and semantic label from the 3DGS model; and the third row shows the loss components for the RGB, depth, and semantic components.
Finally, the fourth row shows the accumulation obtained from the 3DGS model with the empty and dense loss components.

\section{Results}
\label{sec:results}
This section evaluates the key components of our mapping pipeline, beginning with the perception front-end. We first present a quantitative benchmark of various dense depth estimation and semantic segmentation models to assess their performance in the synthetic lunar environment. Subsequently, we analyze the final 3D reconstruction quality, evaluating how the selected perception models impact the geometric accuracy and semantic fidelity of the map. Both quantitative metrics and qualitative results are provided to support the analysis.
The experiments were performed on a system equipped with an AMD Ryzen 9 9950X CPU (16 cores, 5.7\,GHz), 128\,GB RAM, and an NVIDIA GeForce RTX 5090 GPU with 32\,GB VRAM.
All results related to computational requirements (such as memory usage and runtime) are only indicative, as the current implementation, methods, and models could be optimized further.

\subsection{Evaluation Metrics}
This subsection defines the evaluation metrics used to assess the performance of the proposed framework.

\subsubsection{Depth Estimation}
Given an image $I$ and its predicted depth map $\hat{D}$ and a ground truth depth map $D$, we compute the following metrics:
\begin{itemize}
	\item \textbf{Mean Absolute Error (MAE) [m] $\downarrow$}: The average L1 distance between the rendered and ground truth depth.
	      \begin{equation}
		      \text{MAE} = \frac{1}{|I|} \sum_{p \in I} |\hat{D}(p) - D(p)|
	      \end{equation}

	\item \textbf{Absolute Relative Error (AbsRel) $\downarrow$}: The scale-invariant average of the absolute relative difference.
	      \begin{equation}
		      \text{AbsRel} = \frac{1}{|I|} \sum_{p \in I} \frac{|\hat{D}(p) - D(p)|}{D(p)}
	      \end{equation}

	\item \textbf{Threshold Accuracy ($\delta_{25\%}$) $\uparrow$}: The percentage of pixels where the ratio between the rendered and ground truth depth is within a factor of 25\%.
	      \begin{equation}
		      \delta_{25\%} = \frac{1}{|I|} \sum_{p \in I} \mathbbm{1}\left\{\max\left(\frac{\hat{D}(p)}{D(p)}, \frac{D(p)}{\hat{D}(p)}\right) < 1.25\right\}
	      \end{equation}

	\item \textbf{Frames Per Second (FPS) $\uparrow$}: The processing throughput of the depth estimation model.
\end{itemize}

\subsubsection{Semantic Segmentation}
We compute the following metrics over all the pixels in the image for the semantic segmentation models:
\begin{itemize}
	\item \textbf{Intersection over Union (mIoU) $\uparrow$}: The overlap between the predicted and ground truth pixel masks for a single class or label.
	      \begin{equation}
		      \text{IoU} = \frac{\text{TP}}{\text{TP} + \text{FP} + \text{FN}}
	      \end{equation}

	\item \textbf{Accuracy (Acc.) $\uparrow$}: The percentage of all pixels in the image that are correctly classified.
	      \begin{equation}
		      \text{Acc.} = \frac{\text{TP} + \text{TN}}{\text{TP} + \text{TN} + \text{FP} + \text{FN}}
	      \end{equation}
\end{itemize}

\subsubsection{Surface Reconstruction}
To evaluate the final 3D map, we extract a point cloud, $\hat{P}$, by taking the means of the Gaussians in the dense representation and compare it to the ground truth point cloud, $P$. While a more accurate surface reconstruction could be achieved by leveraging the full density field of the Gaussians, this is beyond the scope of our current work. However, since our map is sufficiently dense, using the means of the Gaussians provides a good approximation for evaluation.
\begin{itemize}
	\item \textbf{Accuracy (Chamfer-$L_2$) [cm] $\downarrow$}: The average distance from each point in the reconstructed point cloud to its nearest neighbor in the ground truth point cloud. This measures the correctness of the reconstructed surface.
	      \begin{equation}
		      \text{Accuracy} = \frac{1}{|\hat{P}|} \sum_{\hat{p} \in \hat{P}} \min_{p \in P} \|\hat{p} - p\|_2
	      \end{equation}

	\item \textbf{Completeness (Chamfer-$L_2$) [cm] $\downarrow$}: The average distance from each point in the ground truth to its nearest neighbor in the reconstruction. This measures how well the reconstruction covers the ground truth surface.
	      \begin{equation}
		      \text{Completeness} = \frac{1}{|P|} \sum_{p \in P} \min_{p \in \hat{P}} \|p - \hat{p}\|_2
	      \end{equation}

	\item \textbf{Precision ($d$) [\%] $\uparrow$}: The percentage of reconstructed points within a distance threshold $d$ of the ground truth, measuring correctness.
	      \begin{equation}
		      \text{Precision}(d) = \frac{1}{|\hat{P}|} \sum_{p \in \hat{P}} \mathbbm{1}\left\{\min_{p \in P} \|p - \hat{p}\|_2 < d\right\}
	      \end{equation}

	\item \textbf{Recall ($d$) [\%] $\uparrow$}: The percentage of ground truth points that have a reconstructed point within a distance threshold $d$, measuring completeness.
	      \begin{equation}
		      \text{Recall}(d) = \frac{1}{|P|} \sum_{p \in P} \mathbbm{1}\left\{\min_{p \in \hat{P}} \|p - \hat{p}\|_2 < d\right\}
	      \end{equation}

	\item \textbf{F1-Score ($d$) [\%] $\uparrow$}: The harmonic mean of Precision and Recall, providing a single metric that balances correctness and completeness.
	      \begin{equation}
		      F_1(d) = 2 \cdot \frac{\text{Precision}(d) \cdot \text{Recall}(d)}{\text{Precision}(d) + \text{Recall}(d)}
	      \end{equation}
\end{itemize}

\subsection{Depth Estimation}
\begin{figure*}[t!]
	\centering
	\includegraphics[width=\linewidth]{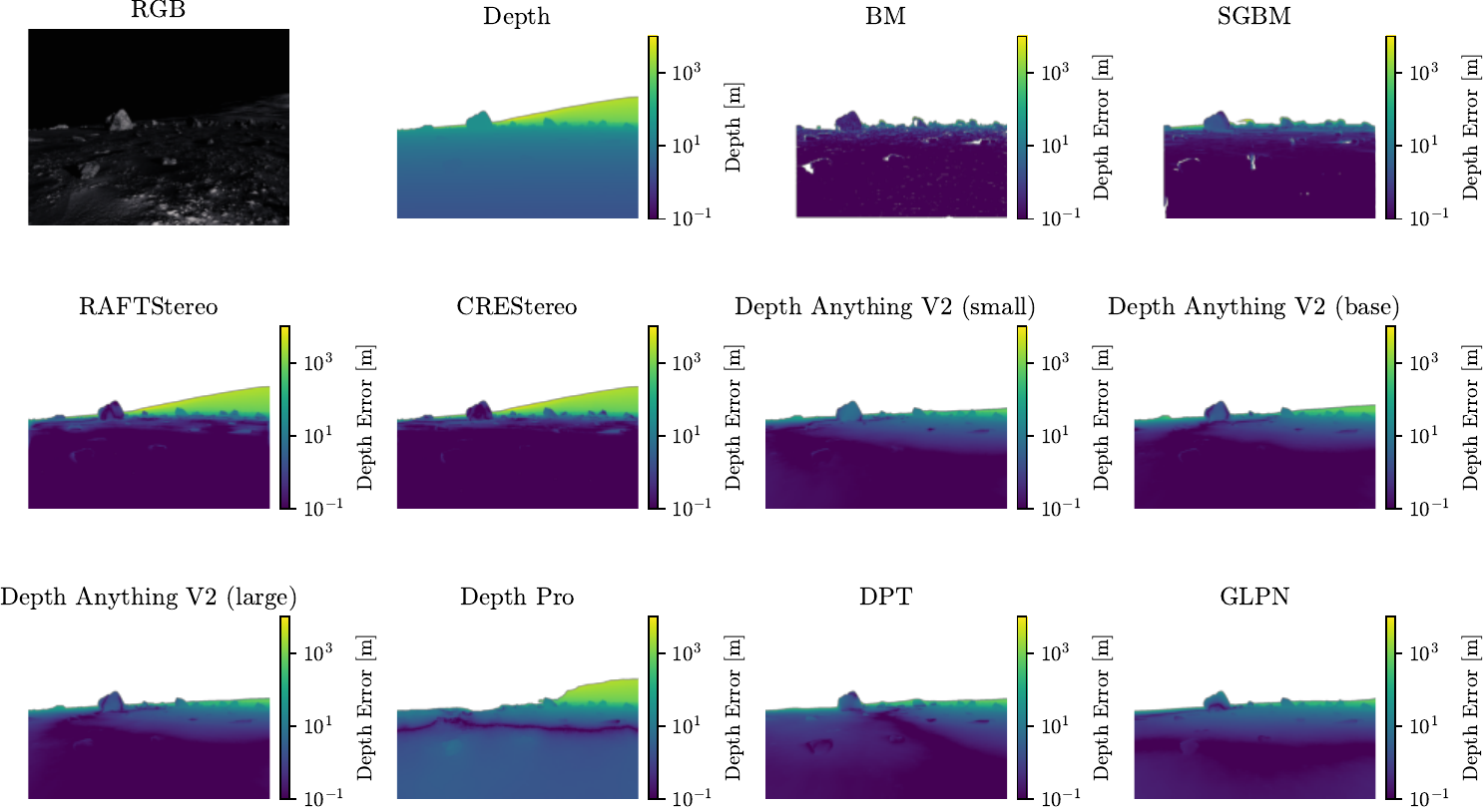}
	\caption{\bfseries Input image and depth estimation errors for a sample of the LuPNT datasets~\cite{vila_lupnt_2025}. The other image used for stereo methods is not shown. The depth and depth errors are in logarithmic scale.}
	\label{fig:depth_errors}
\end{figure*}

This section evaluates the depth estimation models on the spirals LuPNT dataset, combining qualitative analysis with quantitative metrics. \Cref{fig:depth_errors} provides a visual comparison of model outputs, while \Cref{tab:depth_models} lists the numerical results computed for depths up to 20\,m to characterize close-range performance.

Traditional stereo methods, such as Block Matching (BM) and Semi-Global Matching (SGBM), produce depth estimates that have gaps, which are especially notorious when rocks cast long sharp shadows, as they struggle in regions with challenging illumination or low texture, as shown qualitatively in \Cref{fig:depth_errors}. However, they offer high FPS and could also be used in this pipeline. The quantitative results in \Cref{tab:depth_models} reflect this; while their Mean Absolute Error (MAE) is low, this is a consequence of providing depth only in high-confidence areas, as indicated by their low $\delta_{25\%}$.

In contrast, learning-based stereo models like RAFT-Stereo and CREStereo demonstrate superior performance, generating dense depth maps with high completion rates and accuracy. While their MAE can be slightly higher than traditional methods due to denser predictions at long ranges, their low Absolute Relative Error (AbsRel) and significantly higher $\delta_{25\%}$ in \Cref{tab:depth_models} confirm better overall geometric coverage. Notably, RAFT-Stereo achieves better accuracy and higher processing speeds compared to CREStereo, as reported in \Cref{tab:depth_models}. Consequently, we select RAFT-Stereo for our mapping pipeline as it provides the best trade-off between strong accuracy and a frame rate more suitable for real-time operation.

Monocular depth estimation models underperform across all metrics in \Cref{tab:depth_models}, even after being scaled with sparse feature matches. Their performance is likely constrained by the domain shift from the terrestrial datasets they were trained on. Despite their lower accuracy, they remain a potentially valuable source of geometric information when stereo data is unavailable.
It is worth noting that learning-based models were evaluated using weights pretrained on terrestrial datasets. Fine-tuning on domain-specific lunar data would likely mitigate domain shift and improve performance.

\begin{table}[t]
	\centering
	\small
	\caption{\bfseries Comparison of depth estimation models. Monocular depth is scaled using sparse depth estimates from matched features. Metrics computed for a depth of up to 20\,m.}
	\label{tab:depth_models}
	\begin{tabular}{|lrrrrr|}
		\hline
		\textbf{Model}                          &
		\textbf{Param}                          &
		\textbf{MAE}$\downarrow$                &
		\textbf{AbsRel}$\downarrow$             &
		\textbf{$\delta_{25\%}$}$\uparrow$      &
		\textbf{FPS}$\uparrow$                                                                                          \\
		\hline\hline
		\textbf{Stereo}                         &      &               &               &               &                \\
		BM                                      & --   & \textbf{0.04} & 0.01          & 0.38          & \textbf{109.8} \\
		SGBM                                    & --   & 0.09          & 0.01          & 0.39          & 34.7           \\
		RAFTStereo                              & 11M  & 0.05          & \textbf{0.00} & \textbf{0.45} & 4.4            \\
		CREStereo                               & 5M   & 0.09          & 0.01          & \textbf{0.45} & 2.5            \\
		\hline\hline
		\multicolumn{2}{|l}{\textbf{Monocular}} &      &               &               &                                \\
		\multicolumn{2}{|l}{Depth Anything V2}  &      &               &               &                                \\
		~ Small                                 & 25M  & 1.37          & 0.12          & 0.37          & 27.6           \\
		~ Base                                  & 97M  & 1.28          & 0.11          & 0.38          & 20.9           \\
		~ Large                                 & 335M & 1.21          & 0.10          & 0.39          & 12.8           \\
		Depth Pro                               & 61M  & 4.09          & 1.04          & 0.07          & 19.1           \\
		GLPN                                    & 952M & 1.42          & 0.13          & 0.36          & 2.4            \\
		DPT                                     & 343M & 1.41          & 0.14          & 0.36          & 26.0           \\
		\hline
	\end{tabular}
\end{table}

\begin{figure*}[b!]
	\centering
	\includegraphics[width=\linewidth, trim={0 0.5em 0 0.5em}, clip]{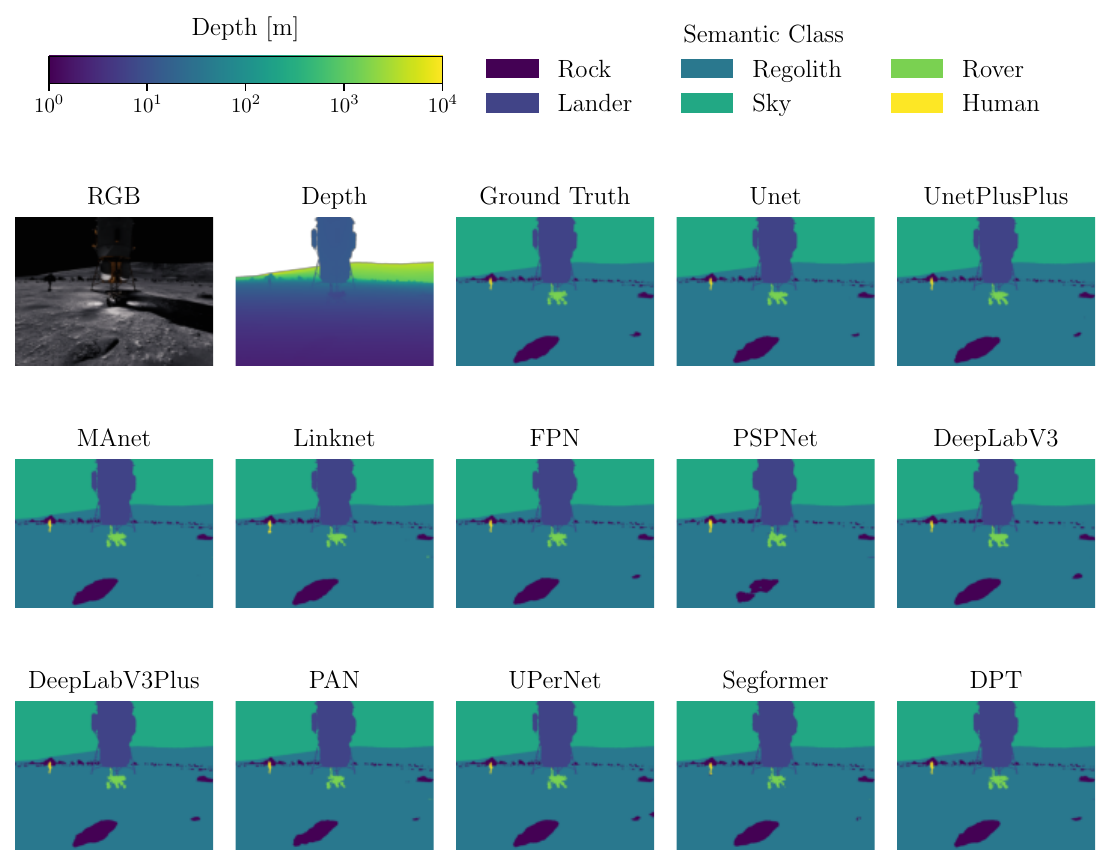}
	\caption{\bfseries Test samples of semantic segmentation predictions for the spirals LuPNT dataset.}
	\label{fig:predictions}
\end{figure*}

\subsection{Semantic Segmentation}
We also benchmarked several semantic segmentation architectures on the spirals LuPNT dataset to evaluate their effectiveness for lunar terrain classification. All models were trained to classify six semantic classes: regolith, rocks, sky, rovers, human, and landers~\cite{vila_lupnt_2025}. The dataset was split into approximately 11,520 images for training, 1,440 for validation, and 1,440 for testing. For training, we used the AdamW optimizer over 40 epochs with a batch size of 8 (effective batch size of 32 with 4 gradient accumulation steps). The initial learning rate was $3\times10^{-4}$ with a weight decay of $5\times10^{-4}$. The learning rate was adjusted using a sequential scheduler: a linear warmup for 5 epochs, followed by a Cosine Annealing decay down to a minimum of $1\times10^{-7}$.

\begin{table*}[t!]
	\centering
	\caption{\bfseries Semantic segmentation results on the spirals LuPNT dataset.}
	\label{tab:semantic_segmentation_models}
	\small
	\begin{tabular}{|l rrrrrrrrrrr|}
		\hline
		\multicolumn{1}{|c}{\multirow{2}{*}{\textbf{Model}}}         &
		\multicolumn{1}{c}{\multirow{2}{*}{\textbf{Size [MB]}}}      &
		\multicolumn{1}{c}{\multirow{2}{*}{\textbf{Params}}}         &
		\multicolumn{1}{c}{\multirow{2}{*}{\textbf{FPS $\uparrow$}}} &
		\multicolumn{6}{c}{\textbf{IoU [\%] $\uparrow$}}             &
		\multicolumn{1}{c}{\textbf{Mean}}                            &
		\multicolumn{1}{c|}{\textbf{Mean}}
		\\
		\cmidrule{4-9}
		                                                             &              &               &                &
		\textbf{Rover}                                               &
		\textbf{Human}                                               &
		\textbf{Lander}                                              &
		\textbf{Rock}                                                &
		\textbf{Regolith}                                            &
		\textbf{Sky}                                                 &
		\multicolumn{1}{c}{\textbf{IoU $\uparrow$}}                  &
		\multicolumn{1}{c|}{\textbf{Acc. $\uparrow$}}
		\\
		\hline
		\hline
		U-Net                                                        & 55.0         & 14.3M         & 188.7          & 78.7          & 70.0          & \textbf{96.7} & 59.0          & 98.0          & 99.5          & \textbf{83.6} & \textbf{98.8} \\
		U-Net++                                                      & 61.0         & 16.0M         & 77.4           & \textbf{80.6} & 65.7          & 96.5          & 59.9          & 98.0          & 99.5          & 83.4          & \textbf{98.8} \\
		MANet                                                        & 83.0         & 21.7M         & 153.8          & 77.6          & 66.4          & 96.6          & \textbf{60.3} & \textbf{98.1} & \textbf{99.7} & 83.1          & \textbf{98.8} \\
		Linknet                                                      & 44.0         & 11.7M         & 257.1          & 72.3          & 57.7          & 95.8          & 52.3          & 97.6          & 99.5          & 79.2          & 98.5          \\
		FPN                                                          & 50.0         & 13.0M         & 275.0          & 76.0          & 65.9          & 94.7          & 55.9          & 97.8          & 99.4          & 81.6          & 98.6          \\
		PSPNet                                                       & \textbf{3.0} & \textbf{0.9M} & \textbf{626.3} & 67.0          & 47.1          & 91.6          & 36.6          & 96.6          & 98.4          & 72.9          & 97.8          \\
		DeepLabV3                                                    & 61.0         & 15.9M         & 169.5          & 76.7          & 62.0          & 96.2          & 56.2          & 97.8          & 99.6          & 81.4          & 98.7          \\
		DeepLabV3Plus                                                & 47.0         & 12.3M         & 332.5          & 77.2          & 61.0          & 96.2          & 55.2          & 97.7          & 99.6          & 81.2          & 98.6          \\
		PAN                                                          & 43.0         & 11.4M         & 301.3          & 72.8          & 56.4          & 95.5          & 53.2          & 97.5          & 99.4          & 79.2          & 98.5          \\
		UPerNet                                                      & 74.0         & 19.5M         & 132.7          & 80.5          & \textbf{71.3} & 96.2          & 54.3          & 97.6          & 99.6          & 83.3          & 98.6          \\
		Segformer                                                    & 45.0         & 11.8M         & 216.0          & 69.5          & 62.4          & 95.1          & 51.4          & 97.5          & 99.3          & 79.2          & 98.5          \\
		DPT                                                          & 159.0        & 41.6M         & 68.5           & 79.7          & 62.7          & 96.4          & 58.2          & \textbf{98.1} & 99.6          & 82.5          & \textbf{98.8} \\
		\hline
	\end{tabular}
\end{table*}
\begin{figure*}[t!]
	\centering
	\begin{subfigure}[b]{0.49\linewidth}
		\includegraphics[width=\linewidth]{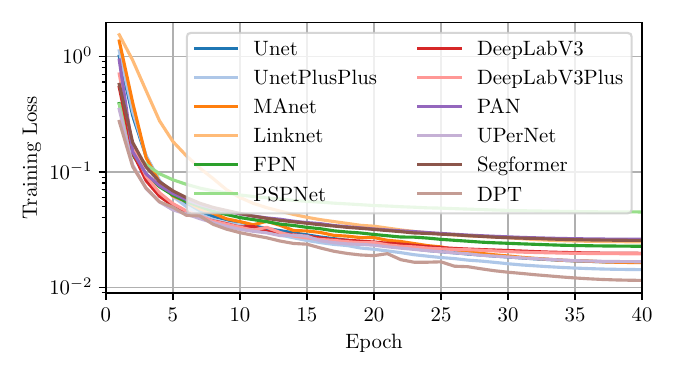}
		\caption{\bfseries Training losses.}
		\label{fig:losses}
	\end{subfigure}
	\begin{subfigure}[b]{0.49\linewidth}
		\includegraphics[width=\linewidth]{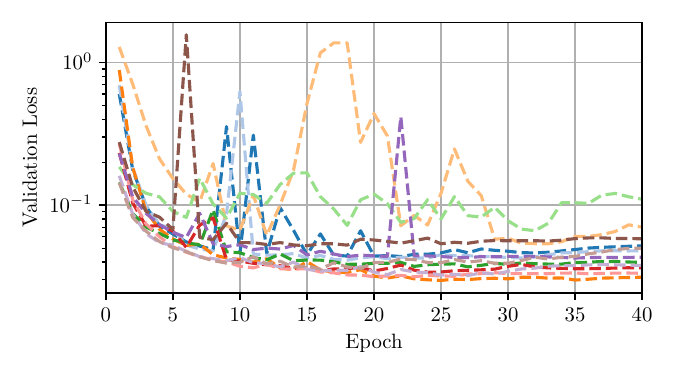}
		\caption{\bfseries Validation losses.}
		\label{fig:val_losses}
	\end{subfigure}
	\caption{\bfseries Training and validation losses for the considered semantic segmentation models.}
\end{figure*}

The loss function $\mathcal{L}$ was a weighted combination of Cross Entropy $\mathcal{L}_{\text{ce}}$, Dice $\mathcal{L}_{\text{dice}}$, and Focal $\mathcal{L}_{\text{focal}}$ losses, formulated as:
\begin{equation}
	\mathcal{L} = \lambda_{ce} \mathcal{L}_{\text{ce}} + \lambda_{dice} \mathcal{L}_{\text{dice}} + \lambda_{focal} \mathcal{L}_{\text{focal}}
\end{equation}
where $\lambda_{ce}=0.2$, $\lambda_{dice}=0.4$, and $\lambda_{focal}=0.4$ are the weights for each loss term.
Cross Entropy Loss $\mathcal{L}_{\text{ce}}$ is a standard pixel-wise loss that penalizes classification errors. Dice Loss $\mathcal{L}_{\text{dice}}$ optimizes the overlap between predicted and ground truth regions, making it effective for handling class imbalance. Finally, Focal Loss $\mathcal{L}_{\text{focal}}$ addresses class imbalance by assigning higher weights to hard-to-classify examples, using a focusing parameter $\gamma = 3.0$ and a balancing parameter $\alpha = 0.25$.

As shown by the training and validation loss curves in \Cref{fig:losses,fig:val_losses}, all models converged successfully, although some spikes are observable in the validation losses, particularly for Linknet. Qualitatively, the majority of architectures are able to accurately segment the primary terrain features, with representative examples shown in \Cref{fig:predictions}. However, some models present lower IoU scores, reflecting the difficulty of the dataset where some images are almost fully dark or feature long shadows due to low sun elevation. The quantitative results in \Cref{tab:semantic_segmentation_models} reveal key performance trade-offs. While U-Net and U-Net++ achieve high overall mean IoU, other models show a slight improvement at specific classes; for instance, MANet is the most effective at rock detection. Notably, PSPNet offers a highly efficient solution, achieving the highest FPS with by far the smallest model size, making it a strong candidate for resource-constrained hardware.
For our downstream mapping task, we selected MANet as it achieved the best performance on the rock, regolith, and sky classes while tied for the highest overall accuracy, though other models would also be suitable.


\begin{figure*}[t!]
	\centering
	\begin{subfigure}[b]{0.49\linewidth}
		\includegraphics[width=\linewidth]{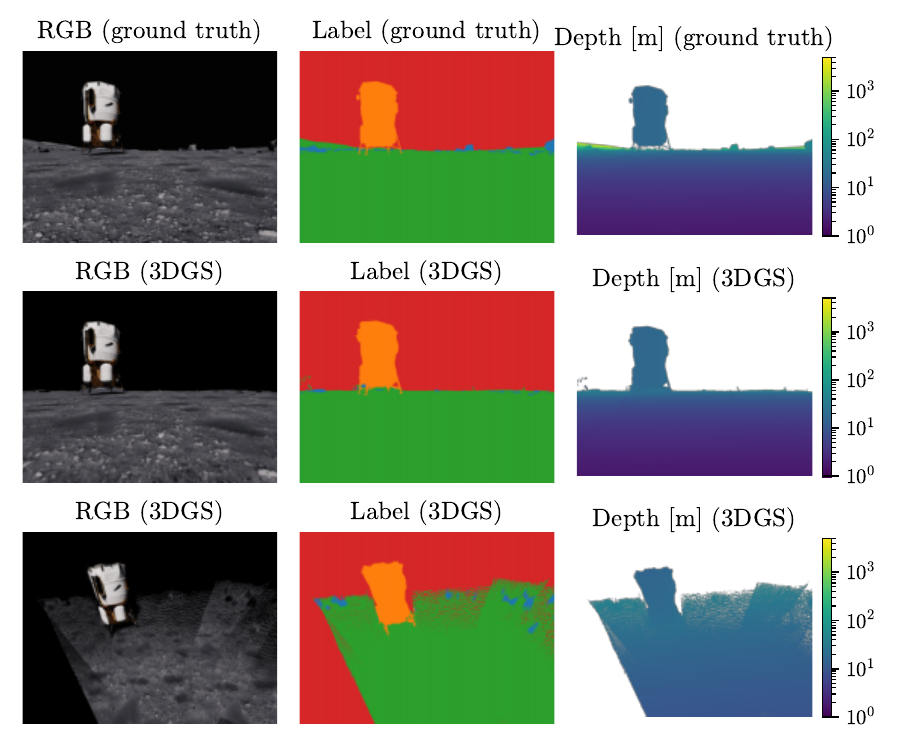}
		\caption{\bfseries Partial.}
	\end{subfigure}
	\hfill
	\begin{subfigure}[b]{0.49\linewidth}
		\includegraphics[width=\linewidth]{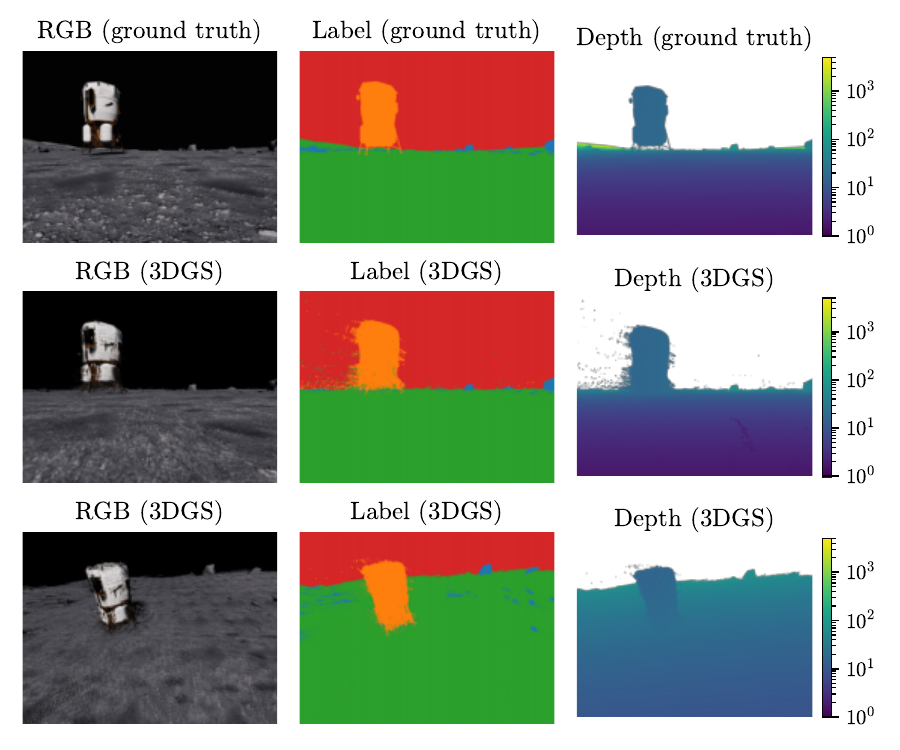}
		\caption{\bfseries Final.}
	\end{subfigure}
	\caption{\bfseries Ground truth and novel views rendered from the 3DGS map.}
	\label{fig:map_3dgs}
\end{figure*}
\begin{figure}[t!]
	\centering
	\includegraphics[width=\linewidth]{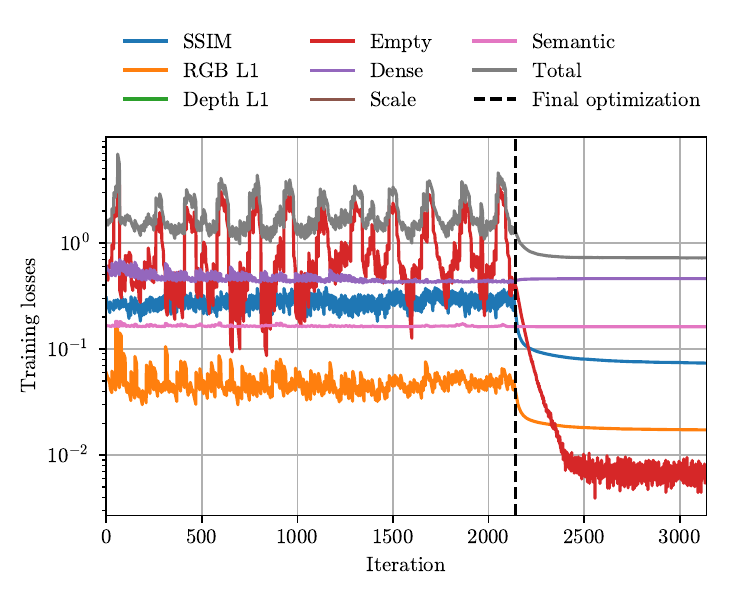}
	\caption{\bfseries 3DGS training losses.}
	\label{fig:all_losses}
\end{figure}
\begin{table*}[t!]
	\centering
	\small
	\caption{\bfseries Summary of the 3DGS map reconstruction results.}
	\label{tab:summary}
	\begin{tabular}[t]{|lrrrrrr|}
		\hline
		\multirow{2}{*}{\textbf{Metric}}               &
		\multicolumn{1}{c}{\textbf{Accuracy}}          &
		\multicolumn{1}{c}{\textbf{Completion}}        &
		\multicolumn{1}{c}{\textbf{Precision}}         &
		\multicolumn{1}{c}{\textbf{Recall}}            &
		\multicolumn{1}{c}{\textbf{F-Score}}           &
		\multicolumn{1}{c|}{\textbf{Height Error}}
		\\
		\multicolumn{1}{|c}{}                          &
		\multicolumn{1}{c}{\textbf{[cm] $\downarrow$}} &
		\multicolumn{1}{c}{\textbf{[cm] $\downarrow$}} &
		\multicolumn{1}{c}{\textbf{[\%] $\uparrow$}}   &
		\multicolumn{1}{c}{\textbf{[\%] $\uparrow$}}   &
		\multicolumn{1}{c}{\textbf{[\%] $\uparrow$}}   &
		\multicolumn{1}{c|}{\textbf{[cm] $\downarrow$}}                                       \\ \hline\hline
		3DGS                                           & 5.4 & 4.8 & 64.3 & 68.6 & 66.4 & 2.8 \\
		+ GT Segmentation                              & 5.7 & 4.8 & 62.0 & 68.5 & 65.1 & 3.0 \\
		+ GT Depth                                     & 4.7 & 4.4 & 69.5 & 74.5 & 71.9 & 2.3 \\
		+ GT Depth + GT Segmentation                   & 4.8 & 4.4 & 69.6 & 74.5 & 72.0 & 2.3 \\
		\hline
		Point Cloud                                    & 9.0 & 4.3 & 47.8 & 68.8 & 56.4 & 3.5 \\
		+ GT Segmentation                              & 8.7 & 4.2 & 47.9 & 68.8 & 56.5 & 3.4 \\
		+ GT Depth                                     & 5.7 & 3.7 & 59.9 & 77.6 & 67.6 & 0.6 \\
		+ GT Depth + GT Segmentation                   & 5.7 & 3.7 & 59.9 & 77.6 & 67.6 & 0.6 \\
		\hline
	\end{tabular}
\end{table*}

\subsection{Surface Reconstruction}
We evaluate our real-time mapping framework on a 120-meter traverse within a scene from the LuPNT trajectories dataset. The input is sourced from a single front-facing stereo camera pair (1024x768 pixel resolution) at a rate of 1 Hz. The pipeline uses MANet for semantic segmentation and RAFT-Stereo for dense depth estimation.
We use a ground truth map sampled at 1 cm resolution.
To benchmark our method and selection of perception models, we compare the reconstruction accuracy of this full pipeline against a configuration that uses ground truth depth and segmentation. For this evaluation, we assume ground truth poses are provided, as our primary focus is to analyze the mapping feasibility and performance independent of a front-end tracking system.

For this preliminary analysis, it is important to note how the surface reconstruction metrics are computed. We extract a point cloud from the 3DGS map by using only the position of each Gaussian. This approach is a simplification, as the center of a Gaussian is not constrained to lie directly on the physical surface. By treating the Gaussians as discrete points, we do not yet leverage the continuous, density-based surface representation that is a key advantage of the method~\cite{wolf_gs2mesh_2025}. We are currently developing a more accurate surface reconstruction pipeline that will use this density information to provide a better evaluation of the map's quality.

\Cref{fig:map_3dgs} shows the output of the mapping process, displaying a partial 3DGS map with the semantic information embedded within each Gaussian, as well as the final, complete map for the entire traverse.
These images are rendered from viewpoints that were not seen during the mapping process, highlighting the framework's ability to build a coherent and renderable 3D model of the environment.
The quality of the scene representation is highlighted in the third row of \Cref{fig:map_3dgs}, which presents a viewpoint above the rover trajectory.

\Cref{fig:all_losses} shows the evolution of the training losses during the incremental mapping process. The losses oscillate as new keyframes are added, introducing new geometry and appearance information, and are subsequently reduced by the online optimization. The frequency and duration of this online optimization are flexible and can be adapted based on the available computational resources on the rover. In the extreme case where no optimization is performed, the method effectively reduces to accumulating 3D points derived from the dense depth and semantic segmentation inputs. In our experiments, we also perform a final optimization step after the traverse is complete to further refine the global map quality.

Although not explored in this work, the incremental generation of these partial maps supports real-time trajectory optimization and path planning. By providing a semantically-aware representation of the environment, the framework allows the rover to refine its path based on the reconstructed geometry and class-specific traversability constraints.
This incremental capability stands in contrast to many existing reconstruction methods that operate in an offline batch mode, assuming that all images and camera pose estimates are available before the mapping process begins.

\begin{figure*}[t!]
	\centering
	\begin{subfigure}[b]{0.32\linewidth}
		\includegraphics[width=\linewidth]{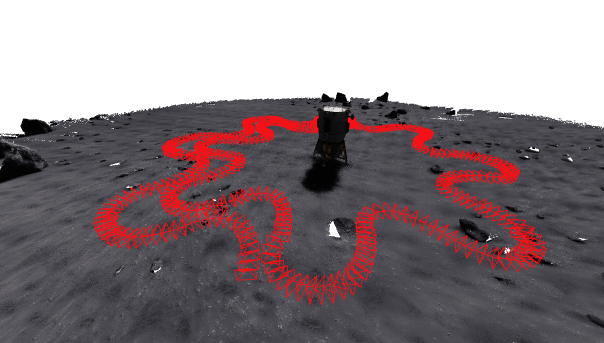}
		\caption{\bfseries Ground truth.}
	\end{subfigure}
	\hfill
	\begin{subfigure}[b]{0.32\linewidth}
		\includegraphics[width=\linewidth]{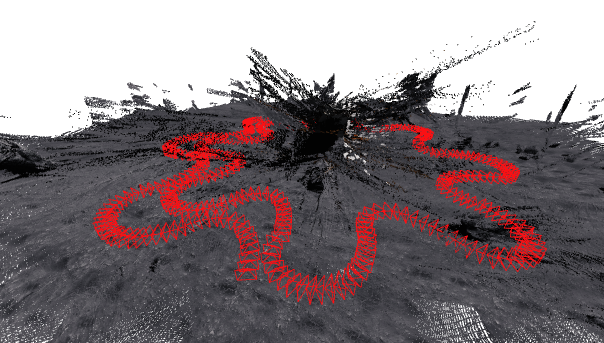}
		\caption{\bfseries Point cloud.}
	\end{subfigure}
	\hfill
	\begin{subfigure}[b]{0.32\linewidth}
		\includegraphics[width=\linewidth]{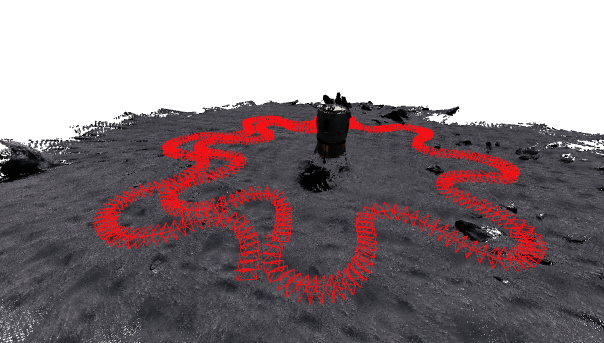}
		\caption{\bfseries 3DGS.}
	\end{subfigure}
	\caption{\bfseries Final map comparison with camera poses used for mapping in red.}
	\label{fig:map_comparison}
\end{figure*}
\begin{table*}[t!]
	\centering
	\small
	\begin{minipage}[t]{0.33\linewidth}
		\centering
		\caption{3DGS memory usage.}
		\label{tab:memory_usage_3dgs}
		\centering
		\begin{tabular}[t]{|lr|}
			\hline
			\textbf{3DGS Model}     & 2375 MB    \\
			\textbf{Num. Gaussians} & 23,944,533 \\
			\hline\hline
			~~Means                 & 274 MB     \\
			~~Rotations             & 365 MB     \\
			~~Scales                & 274 MB     \\
			~~Opacities             & 91 MB      \\
			~~Colors                & 274 MB     \\
			~~Semantics             & 548 MB     \\
			\hline
		\end{tabular}
	\end{minipage}%
	\begin{minipage}[t]{0.33\linewidth}
		\centering
		\caption{Point Cloud memory usage.}
		\label{tab:memory_usage_point_cloud}
		\begin{tabular}[t]{|lr|}
			\hline
			\textbf{Point Cloud} & 331 MB     \\
			\textbf{Num. Points} & 12,406,099 \\
			\hline\hline
			~~Points             & 142 MB     \\
			~~Colors             & 142 MB     \\
			~~Labels             & 47 MB      \\
			\hline
		\end{tabular}
	\end{minipage}
	\begin{minipage}[t]{0.33\linewidth}
		\centering
		\caption{Dataset memory usage.}
		\label{tab:memory_usage_dataset}
		\begin{tabular}[t]{|lr|}
			\hline
			\textbf{Dataset}     & 35,318 MB \\
			\textbf{Num. Frames} & 4,281 MB  \\ \hline\hline
			~~Images             & 19,264 MB \\
			~~Depth Maps         & 12,843 MB \\
			~~Labels             & 3,211 MB  \\
			\hline
		\end{tabular}
	\end{minipage}
\end{table*}

The quantitative results for the map reconstruction are presented in \Cref{tab:summary,tab:surface_reconstruction}. These results show that our full pipeline achieves a mean height error of 2.8 cm. With a 5 cm evaluation threshold, the precision and recall metrics are 64.3\% and 68.6\% respectively, indicating that the majority of the reconstructed surface is geometrically consistent with the ground truth.

A key finding is the minimal impact of the segmentation model's accuracy on the final geometry; using ground truth semantic labels resulted in negligible changes to the aggregate metrics, demonstrating the high performance of the MANet model. The per-class breakdown in \Cref{tab:surface_reconstruction} reveals that reconstruction errors are largest for the rock category (4.0 cm), followed by landers (3.6 cm) and regolith (2.8 cm). This indicates that the reconstruction of small, sharp features like rocks remains the most challenging aspect, likely due to their complex geometry and hard shadows.
A qualitative analysis of the reconstruction shows that the largest geometric errors are concentrated along the dark, high-contrast edges of rocks. These regions represent a common failure mode for both the depth estimation and semantic segmentation models, which struggle to distinguish shadowed rock boundaries from the dark sky or shadowed regolith.

When using estimated depth under ideal pose conditions, the 3DGS pipeline outperforms the point cloud baseline, achieving a height error of 2.8 cm compared to 3.5 cm for the point cloud. This improved accuracy highlights the benefit of the continuous Gaussian representation in smoothing out noise and better approximating the surface geometry. Our next analysis will explore more challenging scenarios, such as re-mapping the same area to evaluate consistency or introducing pose noise. We hypothesize that in such cases, the 3DGS framework's ability to jointly optimize both the map and camera poses to achieve multi-view consistency would demonstrate a significant advantage.

\Cref{tab:memory_usage_3dgs} and \Cref{tab:memory_usage_point_cloud} show the memory requirements for the 120-meter traverse, resulting in a map of approximately 24 million Gaussians and 12 million points. It is important to note that the current memory usage can be significantly reduced, as techniques for pruning and compressing the 3DGS map and baseline point cloud are left as future work~\cite{bagdasarian_3dgszip_2025}. The 3DGS map representation itself (2.4 GB) is significantly more compact than the raw dataset images (19.2 GB), indicating that the 3DGS map effectively compresses the visual information while retaining rendering capabilities. Compared to a traditional semantic point cloud of similar scale, the 3DGS map requires approximately 7 times more memory (331 MB for the point cloud).

Finally, \Cref{fig:map_comparison} allows for a direct visual comparison between the ground truth map, the point cloud baseline, and our 3DGS reconstruction. The figure clearly demonstrates that the 3DGS optimization effectively cleans up noise and artifacts present in the raw point cloud due to depth errors or inconsistencies at object boundaries, resulting in a cleaner and coherent scene representation. This ability to optimize out incorrect points derived from segmentation and dense depth errors is a key advantage of the volumetric Gaussian representation over simple point clouds.

\begin{table*}[h]
	\centering
	\small
	\caption{\bfseries Surface reconstruction results. Precision and recall using a threshold of 5 cm.}
	\label{tab:surface_reconstruction}
	\begin{tabular}[t]{|lrrrrrr|}
		\hline
		\multirow{2}{*}{\textbf{Metric}}               &
		\multicolumn{1}{c}{\textbf{Accuracy}}          &
		\multicolumn{1}{c}{\textbf{Completion}}        &
		\multicolumn{1}{c}{\textbf{Precision}}         &
		\multicolumn{1}{c}{\textbf{Recall}}            &
		\multicolumn{1}{c}{\textbf{F-Score}}           &
		\multicolumn{1}{c|}{\textbf{Height Error}}
		\\
		\multicolumn{1}{|c}{}                          &
		\multicolumn{1}{c}{\textbf{[cm] $\downarrow$}} &
		\multicolumn{1}{c}{\textbf{[cm] $\downarrow$}} &
		\multicolumn{1}{c}{\textbf{[\%] $\uparrow$}}   &
		\multicolumn{1}{c}{\textbf{[\%] $\uparrow$}}   &
		\multicolumn{1}{c}{\textbf{[\%] $\uparrow$}}   &
		\multicolumn{1}{c|}{\textbf{[cm] $\downarrow$}}                                           \\ \hline\hline
		\multicolumn{7}{|l|}{\textbf{3DGS}}                                                       \\ \hline
		Rock                                           & 12.5  & 15.2 & 52.7 & 43.6 & 47.7 & 4.0  \\
		Regolith                                       & 5.1   & 4.7  & 65.8 & 70.4 & 68.0 & 2.8  \\
		Lander                                         & 16.5  & 10.0 & 13.6 & 18.4 & 15.6 & 3.6  \\
		All                                            & 5.4   & 4.8  & 64.3 & 68.6 & 66.4 & 2.8  \\
		\hline
		\multicolumn{7}{|l|}{\textbf{+ GT Segmentation}}                                          \\ \hline
		Rock                                           & 40.5  & 6.1  & 44.0 & 58.5 & 50.2 & 5.3  \\
		Regolith                                       & 5.3   & 4.6  & 64.1 & 70.2 & 67.0 & 3.0  \\
		Lander                                         & 17.3  & 8.8  & 12.9 & 19.9 & 15.6 & 4.1  \\
		All                                            & 5.7   & 4.8  & 62.0 & 68.5 & 65.1 & 3.0  \\
		\hline
		\multicolumn{7}{|l|}{\textbf{+ GT Depth}}                                                 \\ \hline
		Rock                                           & 5.3   & 9.8  & 66.5 & 61.3 & 63.8 & 2.2  \\
		Regolith                                       & 4.8   & 4.4  & 69.4 & 74.7 & 72.0 & 2.4  \\
		Lander                                         & 4.5   & 5.3  & 63.6 & 64.4 & 64.0 & 1.7  \\
		All                                            & 4.7   & 4.4  & 69.5 & 74.5 & 71.9 & 2.3  \\
		\hline
		\multicolumn{7}{|l|}{\textbf{+ GT Depth + GT Segmentation}}                               \\ \hline
		Rock                                           & 4.5   & 4.4  & 69.3 & 74.5 & 71.8 & 2.0  \\
		Regolith                                       & 4.8   & 4.4  & 69.8 & 74.8 & 72.2 & 2.3  \\
		Lander                                         & 4.6   & 5.2  & 63.0 & 64.7 & 63.8 & 1.7  \\
		All                                            & 4.8   & 4.4  & 69.6 & 74.5 & 72.0 & 2.3  \\
		\hline
		\multicolumn{7}{|l|}{\textbf{Point Cloud}}                                                \\ \hline
		Rock                                           & 116.5 & 5.7  & 18.7 & 56.1 & 28.1 & 11.4 \\
		Regolith                                       & 8.2   & 4.2  & 51.7 & 69.8 & 59.4 & 3.7  \\
		Lander                                         & 27.7  & 6.6  & 12.5 & 36.3 & 18.6 & 5.6  \\
		All                                            & 9.0   & 4.3  & 47.8 & 68.8 & 56.4 & 3.5  \\
		\hline
		\multicolumn{7}{|l|}{\textbf{+ GT Segmentation}}                                          \\ \hline
		Rock                                           & 119.8 & 4.9  & 19.9 & 63.7 & 30.3 & 11.4 \\
		Regolith                                       & 7.8   & 4.2  & 52.0 & 69.8 & 59.6 & 3.4  \\
		Lander                                         & 24.2  & 6.5  & 12.9 & 37.4 & 19.1 & 5.2  \\
		All                                            & 8.7   & 4.2  & 47.9 & 68.8 & 56.5 & 3.4  \\
		\hline
		\multicolumn{7}{|l|}{\textbf{+ GT Depth}}                                                 \\ \hline
		Rock                                           & 46.7  & 4.4  & 50.6 & 70.4 & 58.9 & 4.6  \\
		Regolith                                       & 5.9   & 3.7  & 59.1 & 77.1 & 66.9 & 0.7  \\
		Lander                                         & 8.1   & 3.5  & 69.2 & 83.4 & 75.6 & 2.1  \\
		All                                            & 5.7   & 3.7  & 59.9 & 77.6 & 67.6 & 0.6  \\
		\hline
		\multicolumn{7}{|l|}{\textbf{+ GT Depth + GT Segmentation}}                               \\ \hline
		Rock                                           & 5.2   & 3.5  & 61.9 & 82.8 & 70.9 & 1.7  \\
		Regolith                                       & 5.7   & 3.7  & 59.3 & 77.3 & 67.1 & 0.6  \\
		Lander                                         & 4.4   & 3.5  & 69.4 & 82.9 & 75.6 & 1.7  \\
		All                                            & 5.7   & 3.7  & 59.9 & 77.6 & 67.6 & 0.6  \\
		\hline
	\end{tabular}
\end{table*}
\section{Conclusions}
\label{sec:conclusion}
In this work, we presented and evaluated a real-time framework for creating dense, semantic 3D maps of the lunar surface. Our approach integrates dense perception networks with a 3D Gaussian Splatting (3DGS) representation to address the unique challenges of lunar environments. The primary contributions are the benchmarking of modern perception models for this domain and the demonstration of their integration into an incremental mapping pipeline.

Our experimental evaluation on the LuPNT datasets provided several key findings. We identified RAFT-Stereo as a suitable depth estimation model, offering a robust balance between accuracy and real-time processing speed. For semantic segmentation, MANet was selected for its superior performance in detecting rocks, a critical capability for hazard avoidance; its effectiveness was confirmed by the fact that using ground truth segmentation provided only a small improvement to the final map. When using these models, our pipeline achieved a geometric height accuracy of approximately 3 cm. Under the assumption of perfect poses, the reconstruction quality outperformed a traditional point cloud baseline, with the largest errors occurring along the dark edges of rocks and the lander.

This work establishes a foundation for several avenues of future work. The most critical next step is to integrate a tracking front-end to create a complete Simultaneous Localization and Mapping (SLAM) pipeline. This will remove the reliance on ground truth poses and test the 3DGS representation's ability to jointly refine the map and camera trajectory, which we hypothesize will show a significant advantage over simpler methods. Future work will also focus on density-based techniques for surface extraction from the Gaussians to better capture the continuous geometry. Additionally, future efforts must address the limited computational resources typical of planetary missions, investigating model quantization and hardware acceleration to bridge the gap between our desktop-based prototype and a deployable flight system. We also plan to evaluate alternative neural representations, such as NeRF, 2D Gaussian Splatting (2DGS), and Neural Point Clouds, to assess their trade-offs in reconstruction quality and efficiency. Finally, a more rigorous characterization of the system—including its performance with different camera configurations and effects, additional datasets, and the incorporation of perception uncertainty—is necessary to develop navigation and mapping algorithms that can enable rover autonomy.
\section*{Acknowledgements}

We gratefully acknowledge Blue Origin for funding this project and for valuable discussions that contributed to its development.
This manuscript benefited from the use of AI-based assistants, including Claude Sonnet 4.5 and Gemini 2.5 Pro, which were employed for coding assistance and revising. All final content was reviewed and edited by the authors.

\small
\bibliographystyle{IEEEtran}
\bibliography{refs}

\vspace{10em}
\thebiography

\begin{biographywithpic}
	{Guillem Casadesus Vila}{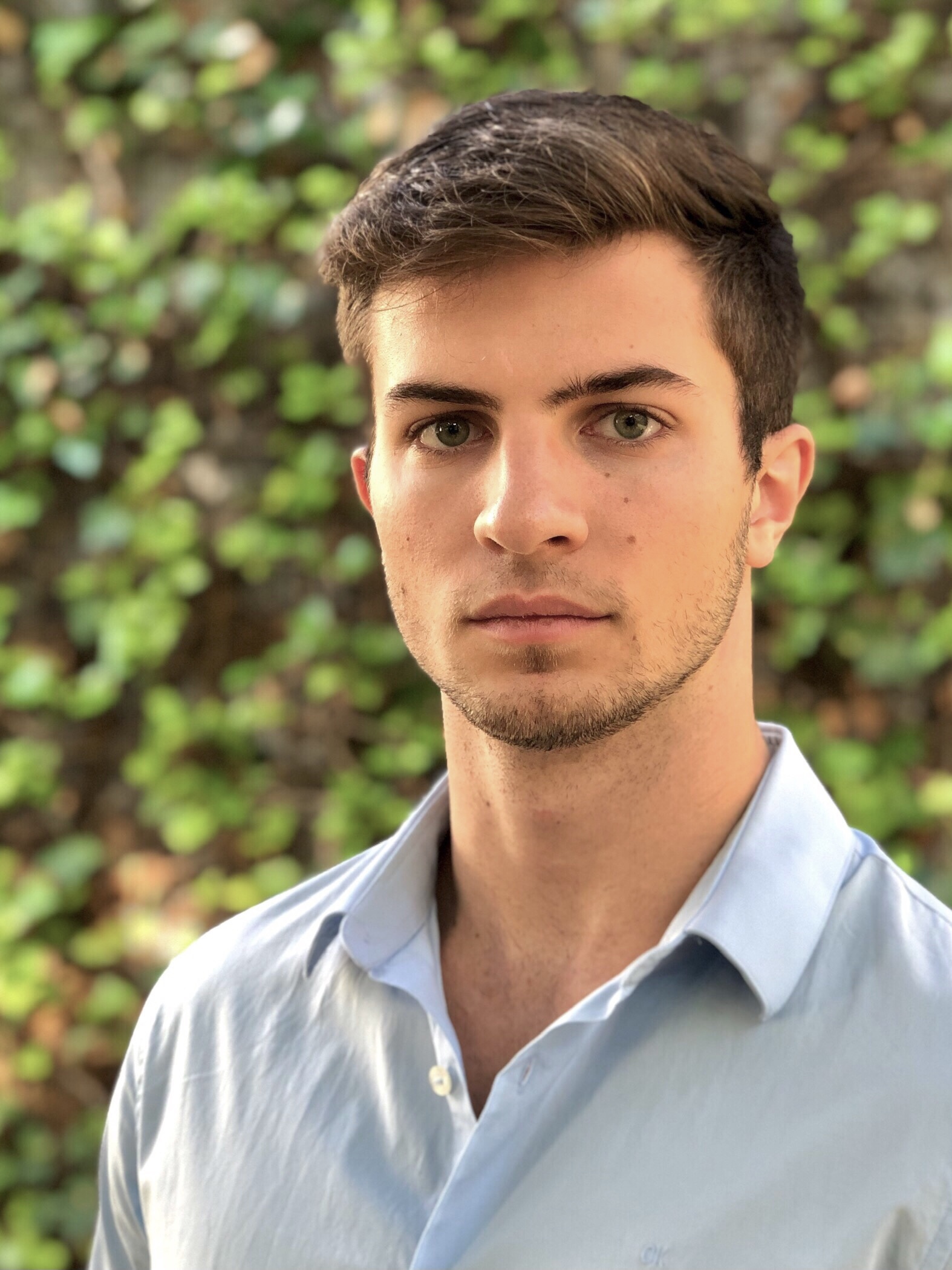}
	is a Ph.D. candidate in the Department of Aeronautics and Astronautics at Stanford University. He received his B.Sc. degree in Aerospace and Telecommunications Engineering in 2022 from the Universitat Politecnica de Catalunya (UPC) under the CFIS program. His research interests include robotic space exploration and space navigation and communication.
\end{biographywithpic}

\begin{biographywithpic}
	{Adam Dai}{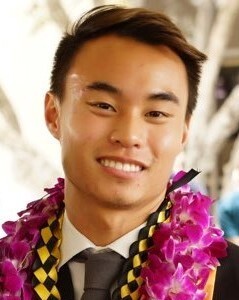}
	is a Ph.D. candidate in the Department of Electrical Engineering at Stanford University. He received his B.Sc. degree in Electrical Engineering with a minor in Computer Science in 2019 from the California Institute of Technology. His research interests include navigation, mapping, and planning in unstructured 3D environments.
\end{biographywithpic}

\begin{biographywithpic}
	{Grace Gao}{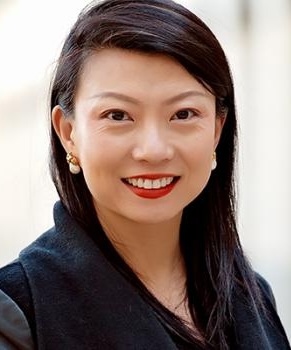}
	is an associate professor in the Department of Aeronautics and Astronautics at Stanford University.
	Before joining Stanford University, she was an assistant professor at University of Illinois at Urbana-Champaign.
	She obtained her Ph.D. degree at Stanford University.
	Her research is on robust and secure positioning, navigation, and timing with applications to manned and unmanned aerial vehicles, autonomous driving cars, as well as space robotics.
\end{biographywithpic}
\vfill

\end{document}